\documentclass[10pt,journal,compsoc]{IEEEtran}

\ifCLASSOPTIONcompsoc
  \usepackage[nocompress]{cite}
\else
  \usepackage{cite}
\fi

\usepackage[pagebackref,breaklinks,colorlinks]{hyperref}
\usepackage{hyperref}
\hypersetup{
    colorlinks=true,
    linkcolor=blue,
    filecolor=blue,      
    urlcolor=blue,
    citecolor=cyan,
}

\usepackage{graphicx}

\usepackage[table,xcdraw,dvipsnames]{xcolor}

\usepackage{bbding}
\usepackage{tikz}
\usepackage{comment}
\usepackage{color}
\usepackage{graphicx}
\usepackage{comment}
\usepackage{amsmath,amssymb}
\usepackage{multirow}
\usepackage{array}
\usepackage{wrapfig}
\usepackage{subfig}

\usepackage[T1]{fontenc}
\usepackage[utf8]{inputenc}
\usepackage{babel}
\usepackage[font=small,labelfont=bf]{caption}
\usepackage{makecell}
\usepackage{booktabs}
\usepackage{enumitem}

\usepackage{float}

\ifCLASSINFOpdf
\else
\fi

\begin{document}
%
\title{Detecting and Grounding Multi-Modal Media Manipulation and Beyond}

\author{Rui Shao,
        Tianxing Wu,
        Jianlong Wu,
        Liqiang Nie,
        Ziwei Liu
\IEEEcompsocitemizethanks{\IEEEcompsocthanksitem Rui Shao, Jianlong Wu, and Liqiang Nie are with School of Computer Science and Technology, Harbin Institute of Technology (Shenzhen) \\
E-mail: shaorui@hit.edu.cn,wujianlong@hit.edu.cn,nieliqiang@gmail.com
\IEEEcompsocthanksitem Tianxing Wu and Ziwei Liu are with the S-Lab, Nanyang Technological University \\
E-mail: twu012@ntu.edu.sg,ziwei.liu@ntu.edu.sg}}


\markboth{Journal of \LaTeX\ Class Files,~Vol.~14, No.~8, August~2015}%
{Shell \MakeLowercase{\textit{et al.}}: Bare Demo of IEEEtran.cls for Computer Society Journals}


\IEEEtitleabstractindextext{%
\begin{abstract}
Misinformation has become a pressing issue. Fake media, in both visual and textual forms, is widespread on the web. 
While various deepfake detection and text fake news detection methods have been proposed, they are only designed for single-modality forgery based on binary classification, let alone analyzing and reasoning subtle forgery traces across different modalities. In this paper, we highlight a new research problem for multi-modal fake media, namely \textbf{D}etecting and \textbf{G}rounding \textbf{M}ulti-\textbf{M}odal \textbf{M}edia \textbf{M}anipulation (\textbf{DGM$^4$}). \textbf{DGM$^4$} aims to not only detect the authenticity of multi-modal media, but also ground the manipulated content (\textit{i.e.,} image bounding boxes and text tokens), which requires deeper reasoning of multi-modal media manipulation. To support a large-scale investigation, we construct the first \textbf{DGM$^4$} dataset, where image-text pairs are manipulated by various approaches, with rich annotation of diverse manipulations. Moreover, we propose a novel \textbf{H}ier\textbf{A}rchical \textbf{M}ulti-modal \textbf{M}anipulation r\textbf{E}asoning t\textbf{R}ansformer (\textbf{HAMMER}) to fully capture the fine-grained interaction between different modalities. \textbf{HAMMER} performs \textbf{1)} manipulation-aware contrastive learning between two uni-modal encoders as shallow manipulation reasoning, and \textbf{2)} modality-aware cross-attention by multi-modal aggregator as deep manipulation reasoning. Dedicated manipulation detection and grounding heads are integrated from shallow to deep levels based on the interacted multi-modal information. \textcolor{black}{To exploit more fine-grained contrastive learning for cross-modal semantic alignment, we further integrate Manipulation-Aware Contrastive Loss with Local View and construct a more advanced model \textbf{HAMMER++}}. Finally, we build an extensive benchmark and set up rigorous evaluation metrics for this new research problem. Comprehensive experiments demonstrate the superiority of \textbf{HAMMER} and \textbf{HAMMER++}; several valuable observations are also revealed to facilitate future research in multi-modal media manipulation.
\end{abstract}

}

\maketitle



\IEEEdisplaynontitleabstractindextext

\IEEEpeerreviewmaketitle

\section{Introduction}
\begin{figure*}[t] 
\begin{center}
  \centering
  \includegraphics[width=0.9\linewidth]{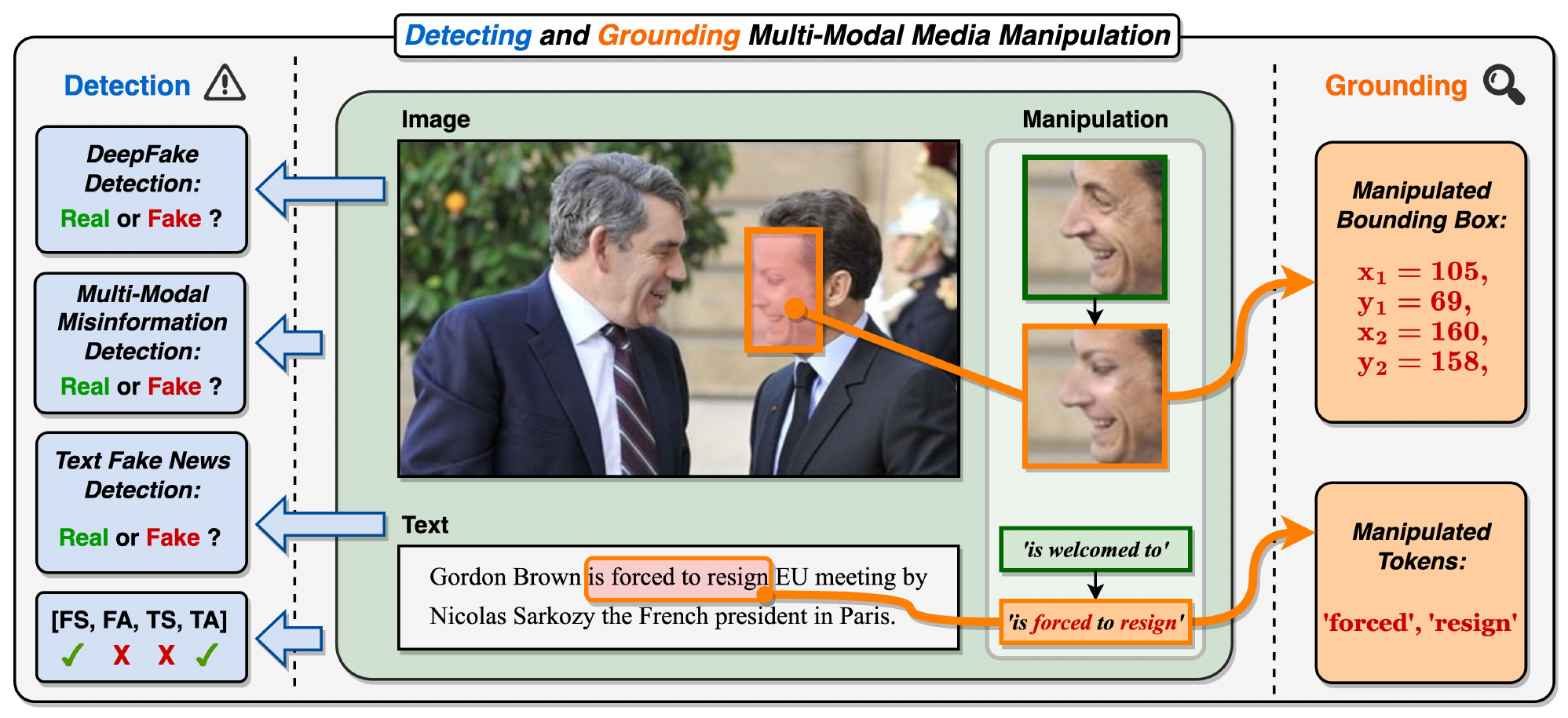}
    \captionof{figure}{Different from existing single-modal forgery detection tasks, \textbf{DGM$^4$} not only performs real/fake classification on the input image-text pair, but also attempts to detect more fine-grained manipulation types and ground manipulated image bboxes and text tokens. They provide more comprehensive interpretation and deeper understanding about manipulation detection besides the binary classification. (FS: Face Swap Manipulation, FA: Face Attribute Manipulation, TS: Text Swap Manipulation, TA: Text Attribute Manipulation)}
  \label{fig:intro}
 \end{center}
\end{figure*}
\begin{table*}[t]
\scriptsize
\centering
\caption{Comparison of the proposed \textbf{DGM$^4$} with existing tasks related to image and text forgery detection.}
\begin{tabular}{lccccc}
\toprule[1pt]
\multicolumn{1}{c}{\multirow{2}{*}{\textbf{Problem Setting}}} & \multicolumn{2}{c}{\textbf{Image Forgery}}                    & \multicolumn{2}{c}{\textbf{Text Forgery}}                     & \multirow{2}{*}{\textbf{\begin{tabular}[c]{@{}c@{}}Multi-Modal\\ Forgery Detection\end{tabular}}} \\ \cline{2-5}
\multicolumn{1}{c}{}                                          & \textbf{Detection}            & \textbf{Grounding}            & \textbf{Detection}            & \textbf{Grounding}            &                                                                                                   \\ \hline
DeepFake Detection~\cite{luo2021generalizing,zhao2021multi}                                            & \CheckmarkBold &  \XSolidBrush  &  \XSolidBrush  &  \XSolidBrush  &     \XSolidBrush                                                                   \\
Text Fake News Detection~\cite{wang2017liar,zellers2019defending}                                     & \XSolidBrush   &  \XSolidBrush  & \CheckmarkBold &   \XSolidBrush &  \XSolidBrush                                                                      \\
Multi-Modal Misinformation Detection~\cite{abdelnabi2022open,luo2021newsclippings}                         & \XSolidBrush   & \XSolidBrush   & \XSolidBrush   & \XSolidBrush   & \CheckmarkBold                                                                     \\
\rowcolor[HTML]{E3DCDC} \textbf{DGM$^4$}        & \CheckmarkBold & \CheckmarkBold & \CheckmarkBold & \CheckmarkBold & \CheckmarkBold                                                                     \\ \bottomrule[1pt]
\end{tabular}
\label{tab:intro}
\end{table*}
With recent advances in deep generative models, increasing hyper-realistic face images or videos can be automatically generated, which results in serious \textit{deepfake} problem~\cite{rossler2019faceforensics++,li2019celeb,dolhansky2020deepfake,jiang2020deeperforensics} spreading massive fabrication on visual media. This threat draws great attention in computer vision community and various deepfake detection methods have been proposed. With the advent of Large Language Model, \textit{e.g.}, BERT~\cite{devlin2019bert}, GPT~\cite{radford2019language}, enormous \textit{text fake news}~\cite{wang2017liar,zellers2019defending} can be readily generated to maliciously broadcast misleading information on textual media. Natural Language Processing (NLP) field pays great attention to this issue and presents diverse text fake news detection methods.

Compared to a single modality, the multi-modal media (in form of image-text pairs) disseminates broader information with greater impact in our daily life. Thus, multi-modal forgery media tends to be more harmful. To cope with this new threat with a more explainable and interpretable solution, this paper proposes a novel research problem, namely \textbf{D}etecting and \textbf{G}rounding \textbf{M}ulti-\textbf{M}odal \textbf{M}edia \textbf{M}anipulation (\textbf{DGM$^4$}). As shown in Table~\ref{tab:intro} and Fig.~\ref{fig:intro}, two challenges are brought by \textbf{DGM$^4$}: \textbf{1)} while current deepfake detection and text fake news detection methods are designed to detect forgeries of single modality, \textbf{DGM$^4$} demands simultaneously detecting the existence of forgery in both image and text modality and \textbf{2)} apart from binary classification like current single-modal forgery detection, \textbf{DGM$^4$} further takes grounding manipulated image bounding boxes (bboxes) and text tokens into account. This means existing single-modal methods are unavailable for this novel research problem. A more comprehensive and deeper reasoning of the manipulation characteristics between two modalities is of necessity. Note that some multi-modal misinformation works~\cite{abdelnabi2022open,luo2021newsclippings} are developed. But they are only required to determine binary classes of multi-modal media, let alone manipulation grounding.

To facilitate the study of \textbf{DGM$^4$}, this paper contributes the first large-scale \textbf{DGM$^4$} dataset. In this dataset, we study a representative multi-modal media form, \textit{human-centric news}. It usually involves misinformation regarding politicians and celebrities, resulting in serious negative influence. We develop two different image manipulation (\textit{i.e.,} face swap/attribute manipulation) and two text manipulation (\textit{i.e.,} text swap/attribute manipulation) approaches to form the multi-modal media manipulation scenario. Rich annotations are provided for detection and grounding, including binary labels, fine-grained manipulation types, manipulated image bboxes and manipulated text tokens. 

Compared to pristine image-text pairs, manipulated multi-modal media is bound to leave manipulation traces in manipulated image regions and text tokens. All of these traces together alter the cross-modal correlation and thus cause semantic inconsistency between two modalities. Therefore, reasoning semantic correlation between images and texts provides hints for the detection and grounding of multi-modal manipulation. To this end, inspired by existing vision-language representation learning works~\cite{li2021align,radford2021learning,kim2021vilt}, we propose a novel \textbf{H}ier\textbf{A}rchical \textbf{M}ulti-modal \textbf{M}anipulation r\textbf{E}asoning t\textbf{R}ansformer (\textbf{HAMMER}) to tackle \textbf{DGM$^4$}. To fully capture the interaction between images and texts, \textbf{HAMMER} \textbf{1)} aligns image and text embeddings through manipulation-aware contrastive learning between two uni-modal encoders as shallow manipulation reasoning and \textbf{2)} aggregates multi-modal embeddings via modality-aware cross-attention of multi-modal aggregator as deep manipulation reasoning. Based on the interacted multi-modal embeddings in different levels, dedicated manipulation detection and grounding heads are integrated hierarchically to detect binary classes, fine-grained manipulation types, and ground manipulated image bboxes, manipulated text tokens. This hierarchical mechanism contributes to more comprehensive manipulation detection and grounding. \textcolor{black}{To exploit more fine-grained contrastive learning in semantic alignment, we further build HAMMER with global and local views of contrastive learning (\textbf{HAMMER++}), which incorporates Manipulation-Aware Contrastive Loss with Local View in the shallow manipulation reasoning. This encourages model to explicitly and adaptively weigh the importance of each local image region or text token in the cross-modal semantic alignment.} Main contributions of our paper:
\begin{itemize}[leftmargin=*]
\item We introduce a new research problem \textbf{D}etecting and \textbf{G}rounding \textbf{M}ulti-\textbf{M}odal \textbf{M}edia \textbf{M}anipulation (\textbf{DGM$^4$}), with the objective of detecting and grounding manipulations in image-text pairs of human-centric news.
\item We contribute a large-scale \textbf{DGM$^4$} dataset with samples generated by two image manipulation and two text manipulation approaches. Rich annotations are provided for detecting and grounding diverse manipulations.
\item We propose a powerful \textbf{H}ier\textbf{A}rchical \textbf{M}ulti-modal \textbf{M}anipulation r\textbf{E}asoning t\textbf{R}ansformer (\textbf{HAMMER}). \textcolor{black}{To explore more fine-grained cross-modal semantic alignment, we construct a more advanced model \textbf{HAMMER++} to further incorporate Manipulation-Aware Contrastive Loss with Local View in the shallow manipulation reasoning.} A comprehensive benchmark is built based on rigorous evaluation protocols and metrics. Extensive quantitative and qualitative experiments demonstrate their superiority.
\end{itemize}

\textcolor{black}{This paper is a substantial extension of our previous CVPR 2023 work~\cite{shao2023cvpr}. We have made three major improvements in this journal version: \textbf{1)} We provide more details regarding the generation pipeline of \textbf{DGM$^4$} dataset. More comprehensive visualizations of image-text pairs in \textbf{DGM$^4$} dataset are also displayed, conveying a more intuitive sense of the dataset; \textbf{2)} To exploit more fine-grained contrastive learning in semantic alignment between images and texts, we further build HAMMER with global and local views of contrastive learning (\textbf{HAMMER++}), which supplements shallow manipulation reasoning with Manipulation-Aware Contrastive Loss with Local View. This allows our model to explicitly weigh the importance of each local image region or text token in the cross-modal semantic alignment; \textbf{3)} We perform more detailed and comprehensive quantitative and qualitative experiments regarding \textbf{HAMMER} and \textbf{HAMMER++}. Various new insights about multi-modal media manipulation are also revealed in this journal version.}

\textcolor{black}{The remainder of this paper is organized as follows. In Section \ref{sec:related work}, relevant studies including deepfake detection and multi-modal misinformation detection are reviewed. Section \ref{sec:dataset} presents the details of generation pipeline and visualizations of samples of the proposed \textbf{DGM$^4$} dataset. Section \ref{sec:method} introduces the manipulation reasoning process of \textbf{HAMMER} and \textbf{HAMMER++}. Section \ref{sec:experiments} reports the experimental results and analyses. Finally, concluding remarks are provided in Section \ref{sec:conclusion}.}

\section{Related Work}
\label{sec:related work}

\noindent \textbf{DeepFake Detection.}
\textcolor{black}{Current deepfake detection methods are built based on spatial and frequency domains, and thus can be roughly categorized into spatial-based and frequency-based deepfake detection. Most of spatial-based deepfake detection methods focus on exploiting visual cues from spatial domain. Li \textit{et al.}~\cite{li2020face} propose Face X-ray to detect the blending boundary caused by face forgery operations as visual artifacts for deepfake detection. Zhao \textit{et al.}~\cite{zhao2021multi} develop a multi-attentional deepfake detection network to adaptively combine low-level textural features and high-level semantic features. 3D decomposition is introduced by Zhu \textit{et al.}~\cite{zhu2021face} into deepfake detection and a two-stream network is constructed to fuse decomposed features for detection. Zhao \textit{et al.}~\cite{zhao2021learning} devise Pair-wise self-consistency learning (PCL) to detect inconsistent source features within the manipulated images. Haliassos \textit{et al.}~\cite{haliassos2021lips} fine-tune a temporal network pretrained on lipreading to capture inconsistencies in semantically high-level mouth movements. On the other hand, several methods pay attention to the frequency domain for capturing spectrum cues. Works~\cite{durall2019unmasking,dzanic2020fourier} find that there exist distinct spectrum distributions and characteristics between real and fake images in the high-frequency part of Discrete Fourier Transform (DFT). A F$^{3}$-Net~\cite{qian2020thinking} is designed to local frequency statistics based on Discrete Cosine Transform (DCT) for forgery mining. A Spatial-Phase Shallow Learning method~\cite{liu2021spatial} is proposed to integrate spatial image and phase spectrum for detecting up-sampling artifacts. Luo \textit{et al.}~\cite{luo2021generalizing} develops a two-stream model to correlate high-frequency features with regular RGB features for capturing generalizable features. Li \textit{et al.}~\cite{li2021frequency} introduce a frequency-aware discriminative feature learning framework to incorporate metric learning into adaptive frequency features learning for face forgery detection. Most of the above deepfake detection methods only perform binary classification in image media, not to mention manipulation grounding across multi-modalities.}

\textcolor{black}{So far, a series of deepfake datasets have been publicly available, such as FaceForensics++~\cite{rossler2019faceforensics++}, Celeb-DF~\cite{li2020celeb}, DeepFake Detection Challenge (DFDC)~\cite{dolhansky2019deepfake}, and DeeperForensics-1.0 (DF1.0)~\cite{jiang2020deeperforensics}. However, only manipulations in image modality are generated and investigated in existing deepfake datasets.}


\noindent \textbf{Multi-Modal Misinformation Detection.}
Several existing works study the detection of multi-modal misinformation~\cite{khattar2019mvae,jin2017multimodal,aneja2021cosmos,luo2021newsclippings,wang2018eann,abdelnabi2022open}. Some of them deal with a small-scale human-generated multi-modal fake news~\cite{khattar2019mvae,jin2017multimodal,wang2018eann}, while others address out-of-context misinformation where a real image is paired with another swapped text without image and text manipulation~\cite{aneja2021cosmos,luo2021newsclippings,abdelnabi2022open}. All of these methods only perform binary classification based on simple image-text correlation. In contrast, \textbf{DGM$^4$} studies large-scale machine-generated multi-media manipulation, which is closer to broad misinformation on the web in practice. Additionally, \textbf{DGM$^4$} requires not only manipulation detection for binary classification, but also manipulation grounding with more interpretation for multi-modal manipulation.


\begin{figure*}[t] 
	\begin{center}
		\includegraphics[width=1\linewidth]{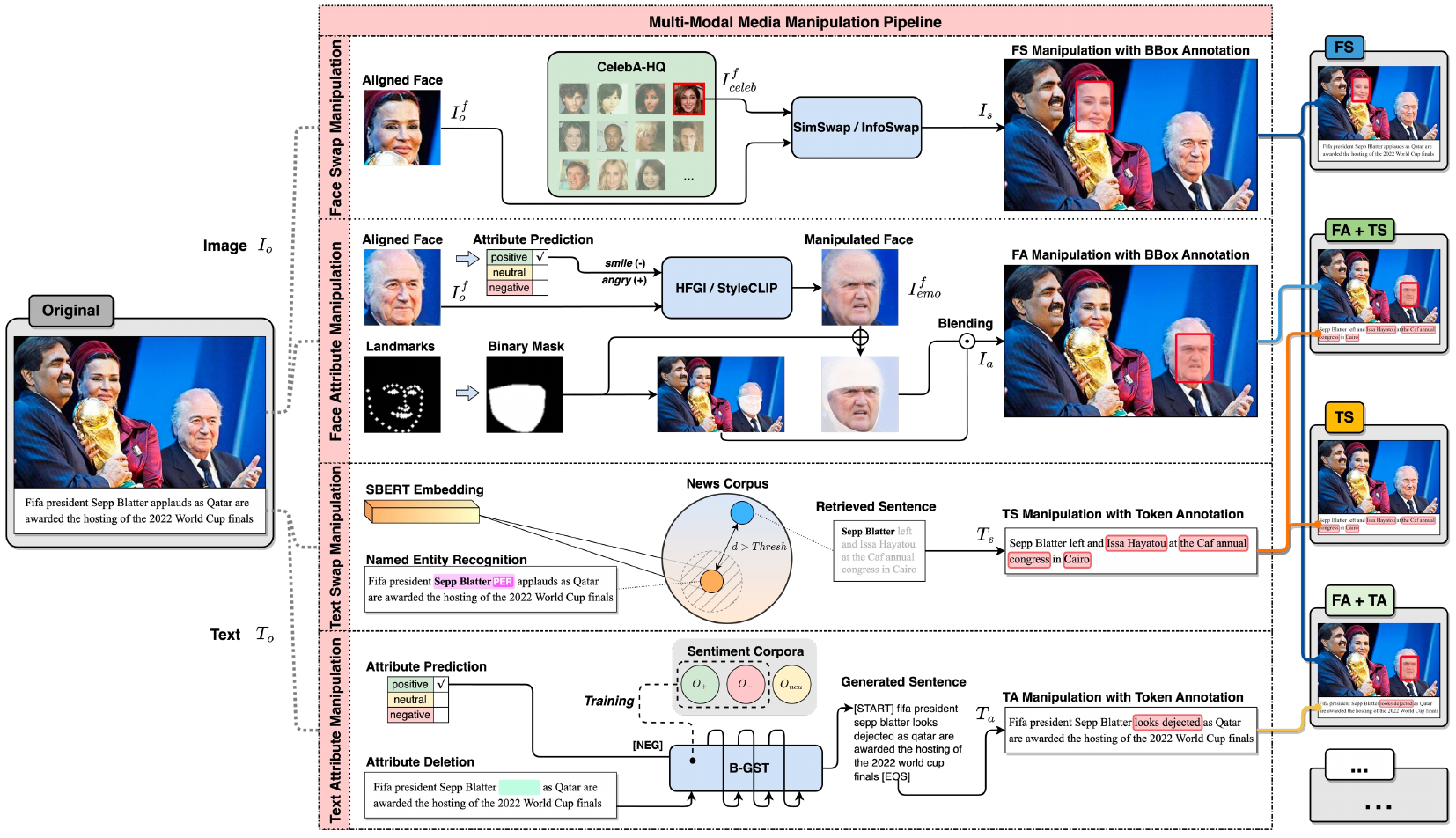}
	\end{center}
	\caption{\textcolor{black}{Generation pipeline of \textbf{DGM$^4$} dataset. An original image-text pair can be manipulated by FS, FA, TS, TA through dedicated manipulation pipelines respectively. The manipulated images and texts are then used to form single-manipulation pairs or are further combined to form more challenging mixed-manipulation pairs.}}
	\label{fig:pipeline}
\end{figure*}

\begin{figure*}[t] 
	\begin{center}
		\includegraphics[width=0.95\linewidth]{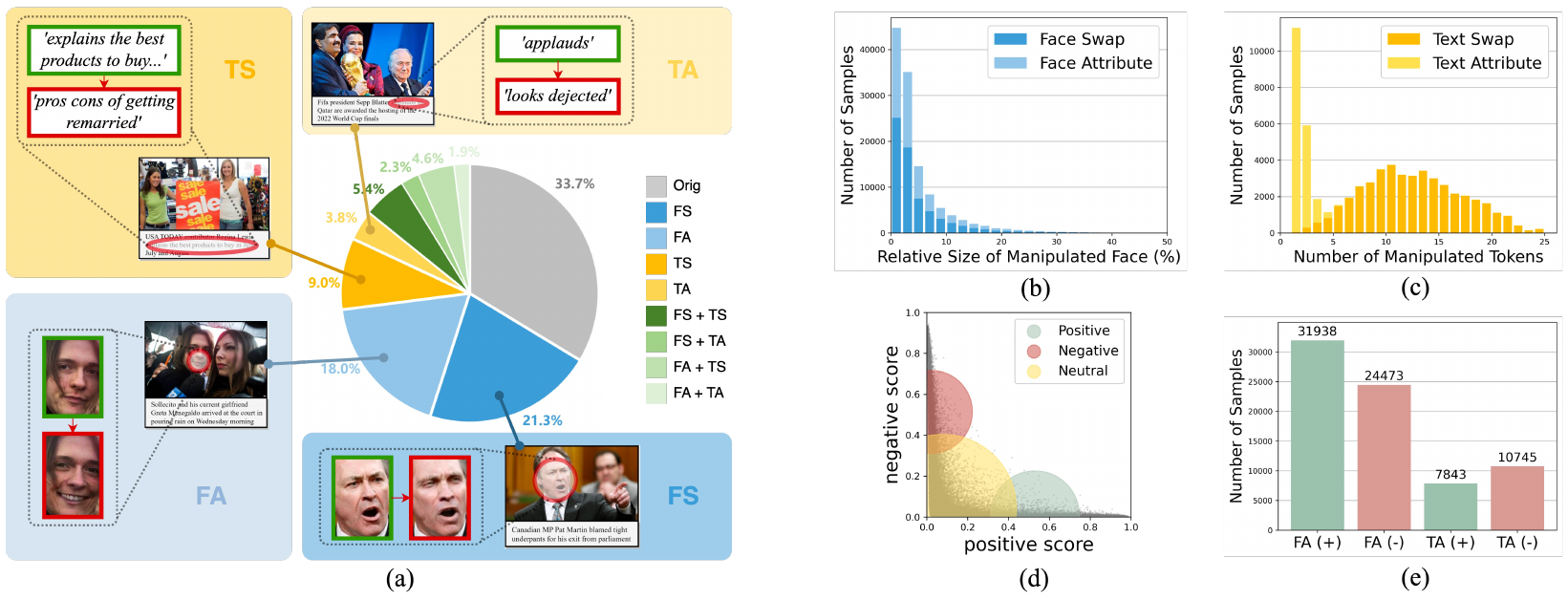}
	\end{center}
	\caption{Statistics of \textbf{DGM$^4$} dataset. (a) Distribution of manipulation classes; (b) manipulated regions of most images are small-size, especially for face attribute manipulation; (c) manipulated tokens of text attribute manipulation are fewer than text swap manipulation; (d) distribution of text sentiment scores in the source pool; (e) number of manipulated samples towards each face/text attribute direction.}
	\label{fig:statistics}
\end{figure*}


\begin{figure*}[!ht]
	\begin{center}
		\includegraphics[width=0.95\linewidth]{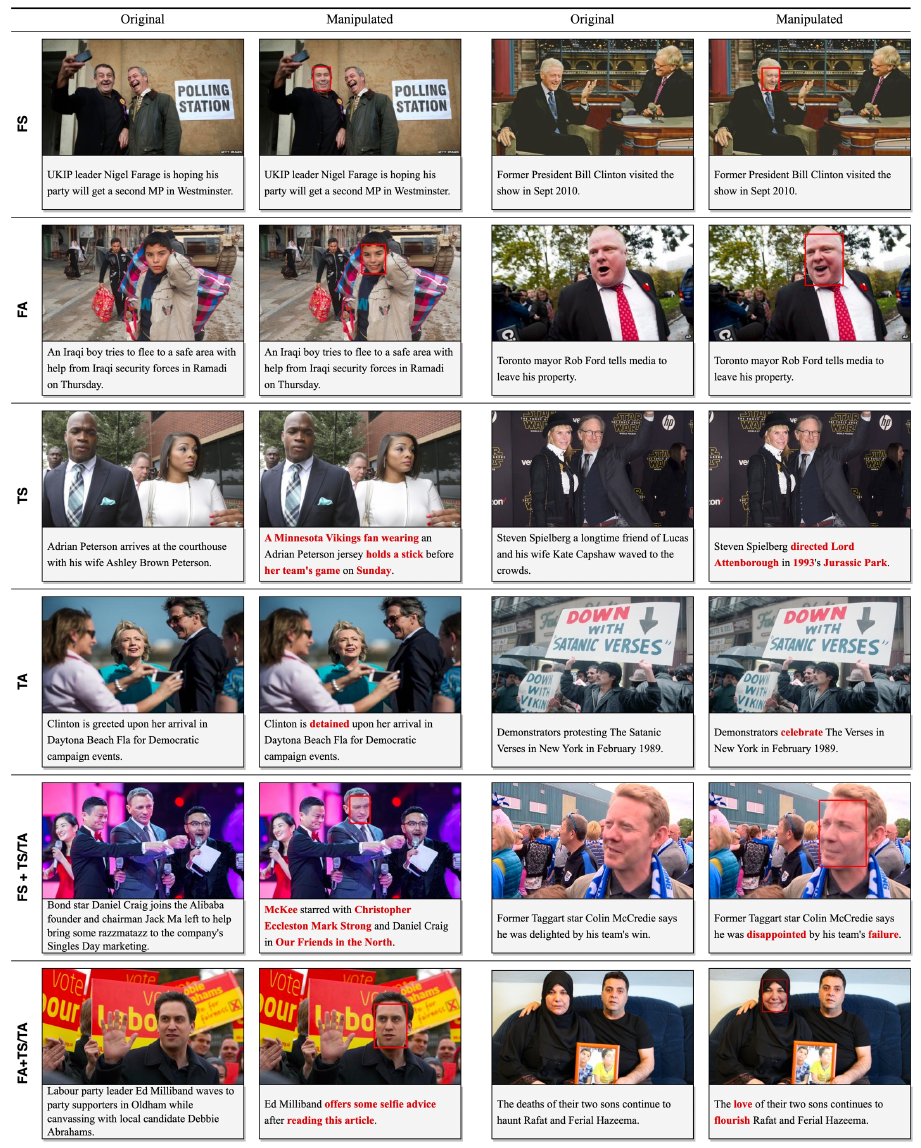}
	\end{center}
	\caption{\textcolor{black}{Manipulation samples based on single-manipulation pairs, \textit{e.g.,} FA, FS, TA, TS, and more challenging mixed-manipulation pairs, \textit{e.g.,} FS+TS/TA and FA+TS/TA, from \textbf{DGM$^4$} dataset.}}
	\label{fig:supp_dataset}
\end{figure*}

\section{Multi-Modal Media Manipulation Dataset}
\label{sec:dataset}
Most of existing misinformation datasets focus on single-modal image forgery ~\cite{rossler2019faceforensics++,li2020celeb,dolhansky2020deepfake,jiang2020deeperforensics} or text forgery ~\cite{wang2017liar,shu2017fake,zellers2019defending}. Some multi-modal datasets are built, but they usually contain a small amount of human-generated fake news~\cite{boididou2018detection,jin2017multimodal} or  out-of-context pairs~\cite{aneja2021cosmos,luo2021newsclippings} for binary forgery detection. To better facilitate the proposed novel research problem, we present \textbf{DGM$^4$} dataset, studying large-scale machine-generated multi-modal media manipulation. \textbf{DGM$^4$} dataset is constructed with diverse manipulation techniques on both \textbf{Image} and \textbf{Text} modality. All samples are annotated with rich, fine-grained labels that enable both \textbf{Detection} and \textbf{Grounding} of media manipulation.

\textcolor{black}{The overall generation pipeline of \textbf{DGM$^4$} dataset is depicted in Fig.~\ref{fig:pipeline}. In this section, we first explain the details regarding the filtering, manipulation and annotation process. Then, some generated fake samples are displayed to provide an intuitive sense of the dataset.}
\subsection{Source Data Collection}
Among all forms of multi-modal media, we specifically focus on \textit{human-centric news}, in consideration of its great public influence. We thus develop our dataset based on the VisualNews dataset~\cite{liu2021visual}, which collects numerous image-text pairs from real-world news sources (The Guardian, BBC, USA TODAY, and The Washington Post). To formulate a human-centric scenario with meaningful context, we further conduct data filtering on both image and text modality, and only keep the appropriate pairs to form the source pool $O=\{p_o|p_o=(I_o, T_o)\}$ for manipulation.
\textcolor{black}{\subsection{Source Data Filtering}}
\textcolor{black}{Before applying manipulations on the source pool $O$, we conduct several data filtering steps, including visual, textual, and cross-modal filtering, to better formulate a human-centric scenario with meaningful context.}

\noindent \textcolor{black}{\noindent \textbf{Visual filtering.} We first use off-the-shelf MTCNN face detector~\cite{zhang2016joint} to: \textbf{1)} filter out images without clearly visible faces and \textbf{2) }restrict the number of faces in each image. This ensures that each image contains human-centric semantics based on clear yet not too complex person-to-person interactions, enabling achievable manipulation reasoning on the dataset.}

\noindent \textcolor{black}{\noindent \textbf{Textual filtering.} Following~\cite{luo2021newsclippings}, we only keep samples with text lengths from 5 to 30 words, guaranteeing the overall quality of captions. We also filter out captions with no verbs to ensure the richness of semantics in texts.}

\noindent \textcolor{black}{\noindent \textbf{Cross-modal filtering.} We observe that some image-text pairs in the original Visual News corpus are essentially not related. Based on this observation, we calculate image-text similarity scores for each pair based on the extracted CLIP~\cite{radford2021learning} image and text features. Pairs with similarity scores lower than 0.2 will be filtered out to make the pristine samples less noisy.}
\subsection{Multi-Modal Media Manipulation}
\label{sec:Dataset_Manipulation}
We employ two types of harmful manipulations on both image and text modality. `Swap' type is designed to include relatively global manipulation traces, while `Attribute' type introduces more fine-grained local manipulations. The manipulated images and texts are then randomly mixed with pristine samples to form a total of 8 fake plus one original manipulation classes. Distribution of manipulation classes and some samples are displayed in Fig.~\ref{fig:statistics} (a). \textcolor{black}{The illustration of whole generation and annotation pipeline can be found in Fig.~\ref{fig:pipeline}.}

\noindent \textbf{Face Swap (FS) Manipulation}. In this manipulation type, the \textit{identity} of the main character is attacked by swapping his/her face with another person. We adopt two representative face swap approaches, SimSwap~\cite{chen2020simswap} and InfoSwap~\cite{gao2021information}. For each original image $I_o$, we choose one of the two approach to swap the largest face $I_o^f$ with a random source face $I_{celeb}^f$ from CelebA-HQ dataset~\cite{karras2018progressive}, producing a face swap manipulation sample $I_{s}$. \textcolor{black}{The quality of all generated samples is examined by checking whether the number of detected faces changes after manipulation (the number of detected faces decreases once the quality of generated images degenerates).}

\textcolor{black}{The MTCNN bbox of the swapped face $y_{\rm box}=\{ x_1, y_1, x_2, y_2\}$ is then saved as annotation for grounding. To obtain this manipulated bbox, we compute the IOU of each MTCNN box with the manipulated area, and select the bbox with maximum IOU to be $y_{\rm box}=\{ x_1, y_1, x_2, y_2\}$. Note that for samples with no image manipulation, we assign the bbox annotation as $y_{\rm box}=\{0,0,0,0\}$.}

\noindent \textbf{Face Attribute (FA) Manipulation}. As a more fine-grained image manipulation scenario, face attribute manipulation attempts to manipulate the \textit{emotion} of the main character's face while preserving the identity. For example, if the original face is smiling, we deliberately edit it to the opposite emotion, \textit{e.g.,} an angry face. To achieve this, we first predict the original facial expression of the aligned face $I_o^f$ with a CNN-based network, then edit the face towards the opposite emotion using GAN-based methods, HFGI~\cite{wang2021HFGI} and StyleCLIP~\cite{patashnik2021styleclip}. After obtaining the manipulated face $I_{emo}^f$, we re-render it back to the original image $I_o$ to obtain the manipulated sample $I_{a}$. Bbox $y_{\rm box}$ is also provided.

\textcolor{black}{The detailed FA pipeline are as follows. \textbf{1)} \textit{Prediction.} For all images in the original source pool $O$, we extract the aligned faces with Dlib toolkit, and predict their emotion labels using VGG19~\cite{simonyan2014very}. Based on the emotion labels, the aligned faces are further classified into `positive' and `negative'. That is, the faces with dominant `happy' emotion are considered as positive, while those with `angry', `disgust', `fear', `sad' or `surprise' are classified as negative. Specifically, faces labeled as `neutral' are classified based on their second dominant emotion labels. In this way, we obtain two emotional face pools $O_+^f$ and $O_-^f$. \textbf{2)} \textit{Manipulation.} The aligned faces are manipulated to their opposite emotion status using GAN-based methods. For HFGI~\cite{wang2021HFGI}, we edit the `smile' attribute of samples in $O_+^f$ to the negative direction, and vice versa. For StyleCLIP~\cite{patashnik2021styleclip}, we similarly edit the faces via the global direction approach with text prompts, such as `an angry face' and `a smiling face'. \textbf{3)} \textit{Blending.} We compute the convex hull of the face area based on face landmarks to obtain a binary mask $I_{m}^f$. Based on the mask, the manipulated face $I_{emo}^f$ is re-rendered back to the original image $I_o$ using Poisson blending, generating the manipulated sample $I_{a}$. The examination of the quality of $I_{a}$ is the same as FS. \textbf{4)} \textit{Annotation.} Identical with the corresponding steps in FS.}

\noindent \textbf{Text Swap (TS) Manipulation}. In this scenario, the text is manipulated by altering its overall \textit{semantic} while preserving words regarding main character. Given an original caption $T_o$, we use Named Entity Recognition (NER) model to extract the person's name as query `PER'. Then we retrieve a different text sample $T_o'$ containing the same `PER' entity from the source corpus $O$. $T_o'$ is then selected as the manipulated text $T_s$. Note that we compute the semantic embedding of each text using Sentence-BERT~\cite{reimers2019sentence} and only accept $T_o'$ that has low cosine similarity with $T_o$. This ensures the retrieved text is not semantically aligned with $T_o$, so that the text semantic regarding the main character in the obtained pair $p_m=(I_o, T_s)$ is manipulated. After that, given $M$ text tokens in $T_s$, we annotate them with a $M$-dimensional one-hot vector $y_{\rm tok} = \{y_i\}^M_{i=1}$, where $y_i \in \{0, 1 \}$ denotes whether the $i$-th token in $T_s$ is manipulated or not.

\textcolor{black}{Implementation details of the above pipeline are as follows. \textbf{1)} \textit{Person entity extraction.} We use the xlm-R-based English 4-class NER model from Flair~\cite{schweter2020flert} to extract the person's names from original text $T_o$. Since there could be more than one person in a piece of news, we only use the first `PER' entity for retrieval. Furthermore, to avoid cases where the query person is mentioned in texts but is not pictured in images, we also filter the entities in possessive form following~\cite{luo2021newsclippings}. \textbf{2)} \textit{Retrieval filtering.} After retrieving $T_o'$ with identical `PER' but distinct sentence similarity, we further extract the CLIP and Place365 image embeddings, and filter out retrieved texts whose paired image $I_o'$ has high CLIP or Place365 similarities (CLIP\textgreater0.9 or Place365\textgreater0.96) with the original image $I_o$. This is because the retrieved text could be describing the \textit{same} news from another aspect even if the sentences are quite different, which would make $\{I_o, T_o'\}$ a plausible pair rather than a forgery. By ensuring $I_o$ and $I_o'$ are semantically distinct, we can finally obtain the text swap manipulation pair $p_m=(I_o, T_s)$. \textbf{3)} \textit{Annotation.} To produce cleaner annotations, we label the manipulated tokens at \textit{phrase level}. Specifically, we use a LSTM-CRF-based chunking model from Flair~\cite{akbik2018coling} to split the original text into phrases, and label all tokens in the non-overlapping phrases as `fake'. To make the annotation more meaningful, we only consider the `NP', `VP', `ADJP' and `ADVP' chunks~\cite{tjong2000introduction}, as they contain most of the semantics. The chunks without essential semantic meaning (\textit{e.g.,} `to', `at', `as well as') are not included in the annotation. Note that we deliberately keep dates and places (\textit{e.g.,} `9th March', `New York City') in annotation, although they are not highly related to the semantics. This is because not only they are important components of news, but also they provide a chance to study deeper manipulation reasoning with fact-checking techniques like knowledge graph in future research.}

\noindent \textbf{Text Attribute (TA) Manipulation}. Although news is a relatively objective media form, we observe that a considerable portion of news samples $p_o \in O$ still carry \textit{sentiment} bias within the text $T_o$, as depicted in Fig.~\ref{fig:statistics} (d). The malicious manipulation of text attributes, especially its sentiment tendency, could be more harmful and also harder to be detected as it causes less cross-modal inconsistency than text swap manipulation. To reflect this specific situation, we propose Text Attribute Manipulation (TA) as another text manipulation type.

\textcolor{black}{Normally, TA can be regarded as a standard style transfer task in NLP. However, existing sentiment transfer methods~\cite{Ke2019controllable, RameshKashyap2022SoDY} and datasets (\textit{e.g.,} Yelp, Amazon~\cite{li2018delete} and IMDB~\cite{dai2019style}) mainly focus on online reviews data, thus are not applicable to human-centric news. To overcome this limitation, we use a RoBERTa-based sentiment analysis model trained on Twitter~\cite{barbieri2020tweeteval} to split our source data into positive, negative and neutral corpora $\{O_+, O_-,  O_{neu}\}$, and drop the neutral corpus. In this way, we acquire a customized sentiment news dataset $\{O_+, O_-\}$ containing 21,194 positives and 19,602 negatives. Based on this dataset, style transfer methods can be deployed to news. Following~\cite{Sudhakar2019TransformingDR}, we first delete all sentiment words from the original text $T_o$ using the BERT-basted Delete Transformer (DT). Then we generate $T_a$ with the remaining content using the GPT-based Blind Generative Style Transformer (B-GST) trained on our sentiment corpora $\{O_+, O_-\}$. After that, we conduct sentiment analysis on the generated sentence $T_a$ to ensure that its sentiment is successfully flipped. Filtering steps are further applied to remove $T_a$ with bad quality, \textit{e.g.,} repeated phrases in a sentence. Finally, we annotate the sentiment words added by B-GST as fake to form a ground-truth vector $y_{\rm tok}$, providing the fine-grained annotation.}

\noindent \textbf{Combination and Perturbation}. Once all single-modal manipulations are finished, we combine the obtained manipulation samples $I_s$, $I_a$, $T_s$ and $T_a$ with the original $(I_o, T_o)$ pairs. This forms a multi-modal manipulated media pool with full manipulation types: $P=\{p_m|p_m=(I_x, T_y), x,y\in\{o, s, a\}\}$. Each pair $p_m$ in the pool is provided with a binary label $y_{\rm bin}$, a fine-grained manipulation type annotation $y_{\rm mul}$, aforementioned annotations $y_{\rm box}$ and $y_{\rm tok}$. $y_{\rm bin}$ describes whether the image-text pair $p_m$ is real or fake, and $y_{\rm mul} = \{y_j\}^4_{j=1}$ is a 4-dimensional vector denoting whether the $j$-th manipulation type (\textit{i.e.,} FS, FA, TS, TA) appears in $p_m$.
\textcolor{black}{To better reflect the real-world situation where manipulation traces may be covered up by noise, we employ random image perturbations on 50\% of the media pool $P$, such as JPEG compression, Gaussian Blur, \textit{etc}, which are randomly added or mixed on 50\% of the samples. These perturbations are very common distortions existing in multi-modal media in the real life and some manipulation artifacts (\textit{e.g.,} over-smooth skin, face boundaries) could be covered by them. The comprehensiveness of perturbations in \textbf{DGM$^4$} dataset ensures its validity and diversity to better simulate multi-modal media manipulation in various real-world scenarios.}\\
\textcolor{black}{\subsection{Dataset Split}}
\textcolor{black}{After obtaining the final multi-modal media manipulation pool $P=\{p_m|p_m=(I_x, T_y), x,y\in\{o, s, a\}\}$, we split the data into training, validation and test sets roughly according to a portion of 7:1:2 based on the sample IDs. This guarantees that different manipulation samples originating from the same $p_o=(I_o, T_o)$ do not appear at the same time between training set and validation/test sets. Note that because of TS, text from an image-text pair in the training set could be swapped to a pair in the test set. To avoid this problem, we further conduct data deduplication to ensure that the validation and test sets only contain non-overlapped text samples.}\\
\subsection{Dataset Statistics}
\label{sec:Dataset_Statistics}
The overall statistics of \textbf{DGM$^4$} dataset are illustrated in Fig.~\ref{fig:statistics} (a). It consists a total of \textbf{230k} news samples, including 77,426 pristine image-text pairs and 152,574 manipulated pairs. The manipulated pairs contain 66,722 face swap manipulations, 56,411 face attribute manipulations, 43,546 text swap manipulations and 18,588 text attribute manipulations. $\sim$1/3 of the manipulated images and $\sim$1/2 of the manipulated text are combined together to form 32,693 mixed-manipulation pairs. Since both image and text attributes can be edited towards two opposite sentiment directions, we keep a balanced proportion to create an emotionally-balanced dataset, as shown in Fig.~\ref{fig:statistics} (e).

Furthermore, it can be observed from Fig.~\ref{fig:statistics} (b)-(c) that the manipulated regions of most images and the number of manipulated text tokens are relatively small. This indicates \textbf{DGM$^4$} dataset provides a much more challenging scenario for forgery detection compared to existing deepfake and multi-modal misinformation datasets. 
\textcolor{black}{\subsection{Dataset Samples}}
\textcolor{black}{We present more manipulation samples from \textbf{DGM$^4$} dataset in Fig.~\ref{fig:supp_dataset}. As can be seen from FA and FS, these manipulation approaches can be deployed in faces with different sizes, generating hyper-realistic swapped faces or edited faces with minor manipulation traces. In particular, it can be observed that FA can apparently alter the emotions of original faces from neutral or anger to happiness.}

\textcolor{black}{In addition, TS swaps the text content except for tokens regarding the main character. This manipulates the semantics of the whole text with respect to the same character. Comparatively, more fine-grained manipulation on single or few tokens are introduced in TA. It can also flip the sentiment of texts by replacing the positive tokens (\textit{e.g.,} `greeted') with negative ones (\textit{e.g.,} `detained'), and vice versa. This smaller manipulation by TA is expected to be harder for detection and grounding than TS, which is also proved later by the statistics regarding multi-label classification in our experiments (Fig.~\ref{fig:multicls}).}

\textcolor{black}{It is natural to combine both image and text manipulations to form more challenging mixed-manipulation pairs like FS+TS/TA and FA+TS/TA, as displayed in the last two rows of Fig.~\ref{fig:supp_dataset}. This demands more advanced and detailed manipulation reasoning and leaves much room for the improvement of multi-modal manipulation detection and grounding.}\\
\begin{figure*}[t] 
	\begin{center}
		\includegraphics[width=1\linewidth]{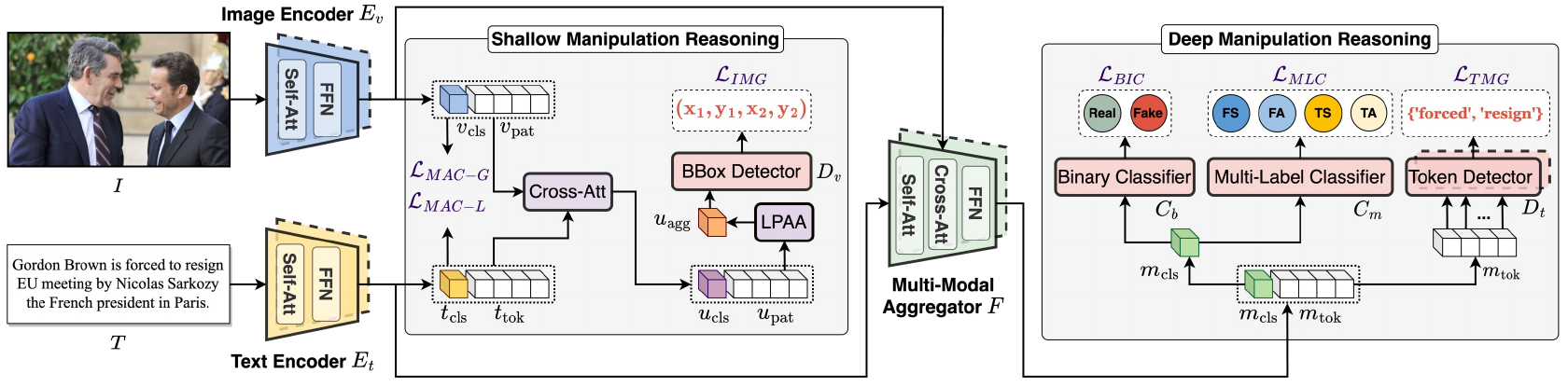}
	\end{center}
	\caption{Overview of proposed \textbf{HAMMER++}. \textcolor{black}{It \textbf{1)} aligns image and text embeddings through manipulation-aware contrastive learning with global and local views} between Image Encoder $E_v$, Text Encoder $E_t$ in shallow manipulation reasoning and \textbf{2)} further aggregates multi-modal embeddings via modality-aware cross-attention of Multi-Modal Aggregator $F$ in deep manipulation reasoning. Based on the interacted multi-modal embeddings in different levels, various manipulation detection and grounding heads (Multi-Label Classifier $C_m$, Binary Classifier $C_b$, BBox Detector $D_v$, and Token Detector $D_t$) are integrated to perform their tasks hierarchically. Modules with dashed lines mean they are the corresponding momentum versions of Image Encoder, Text Encoder, Multi-Modal Aggregator and Token Detector, respectively.}
	\label{fig:framework}
\end{figure*}
\begin{figure}[t] 
	\begin{center}
		\includegraphics[width=1\linewidth]{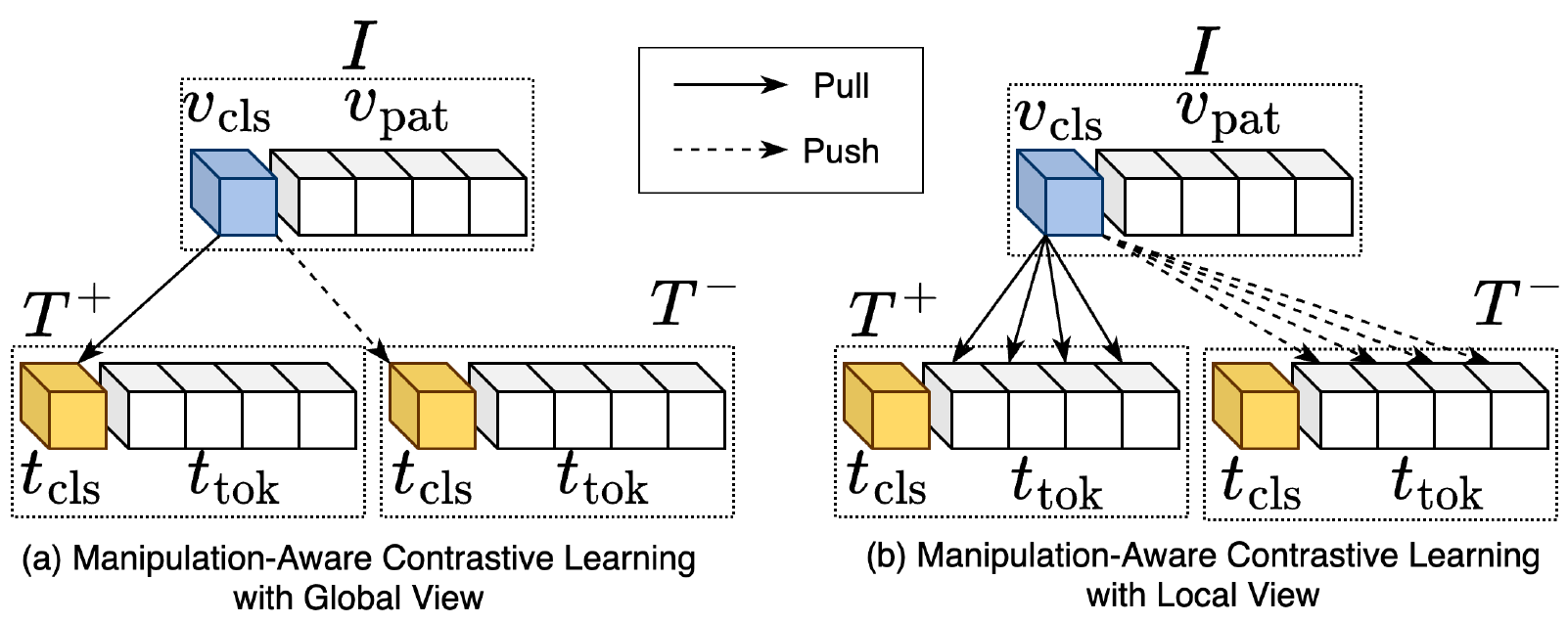}
	\end{center}
	\caption{\textcolor{black}{Comparison between Manipulation-Aware Contrastive with (a) Global View and (b) Local View. Without loss of generality, we use image-to-text contrastive loss as an example.}}
	\label{fig:MAC}
\end{figure}

\section{HAMMER and HAMMER++}
\label{sec:method}
To address \textbf{DGM$^4$}, we propose a \textbf{H}ier\textbf{A}rchical \textbf{M}ulti-modal \textbf{M}anipulation r\textbf{E}asoning t\textbf{R}ansformer (\textbf{HAMMER}), which is composed of two uni-modal encoders (\textit{i.e.,} Image Encoder $E_v$, Text Encoder $E_t$), Multi-Modal Aggregator $F$, and dedicated manipulation detection and grounding heads (\textit{i.e.,}  Binary Classifier $C_b$, Multi-Label Classifier $C_m$, BBox Detector $D_v$, and Token Detector $D_t$). All of these uni-modal encoders and multi-modal aggregator are built based on transformer-based architecture~\cite{vaswani2017attention}. As mentioned above, modeling semantic correlation and capturing semantic inconsistency between two modalities can facilitate detection and grounding of multi-modal manipulation. However, there exist two challenges \textbf{1)} as discussed in Sec.~\ref{sec:Dataset_Statistics} and shown in Fig.~\ref{fig:statistics} (b)-(c), a large portion of multi-modal manipulations are minor and subtle, locating in some small-size faces and a few word tokens and \textbf{2)} much visual and textual noise~\cite{li2021align} exists in multi-modal media on the web. As a result, some semantic inconsistencies caused by manipulation may be neglected or covered by noise. This demands more fine-grained reasoning of multi-modal correlation. To this end, we devise \textbf{HAMMER} to perform hierarchical manipulation reasoning which explores multi-modal interaction from shallow to deep levels, along with hierarchical manipulation detection and grounding. In the shallow manipulation reasoning, we carry out semantic alignment between image and text embeddings through Manipulation-Aware Contrastive Loss with Global View $\mathcal{L}_{MAC-G}$, and conduct manipulated bbox grounding under Image Manipulation Grounding Loss $\mathcal{L}_{IMG}$. In the deep manipulation reasoning, based on deeper interacted multi-modal information generated by Multi-Modal Aggregator, we then detect binary classes with Binary Classification Loss $\mathcal{L}_{BIC}$, fine-grained manipulation types with Multi-Label Classification Loss $\mathcal{L}_{MLC}$, and ground manipulated text tokens via Text Manipulation Grounding Loss $\mathcal{L}_{TMG}$. \textcolor{black}{To exploit more fine-grained contrastive learning for cross-modal semantic alignment, as illustrated in Fig.~\ref{fig:framework}, we construct a more advanced model, \textbf{HAMMER} with global and local views of contrastive learning (\textbf{HAMMER++}), which incorporates Manipulation-Aware Contrastive Loss with Local View $\mathcal{L}_{MAC-L}$ in the shallow manipulation reasoning.} By combing all the above losses, manipulation reasoning is performed hierarchically, contributing to a joint optimization framework for \textbf{HAMMER} and \textbf{HAMMER++} respectively as follows:
\begin{equation}
\footnotesize
\begin{split}
\mathcal{L} = \mathcal{L}_{MAC-G} + \mathcal{L}_{IMG} + \mathcal{L}_{MLC} + \mathcal{L}_{BIC} + \mathcal{L}_{TMG}
\end{split}
\label{equ:opt-HAMMER} 
\end{equation}
\begin{equation}
\footnotesize
\begin{split}
\mathcal{L}_{++} = \mathcal{L}_{MAC-G} + \mathcal{L}_{MAC-L}+ \mathcal{L}_{IMG} + \mathcal{L}_{MLC} + \mathcal{L}_{BIC} + \mathcal{L}_{TMG}
\end{split}
\label{equ:opt-HAMMER++} 
\end{equation}
\subsection{Shallow Manipulation Reasoning}
Given an image-text pair $(I, T) \sim P$, we patchify and encode image $I$ into a sequence of image embeddings via self-attention layers and feed-forward networks in Image Encoder as $E_v(I)=\{v_{\rm cls}, v_{\rm pat} \}$, where $v_{\rm cls}$ is the embedding of \texttt{[CLS]} token, and $v_{\rm pat} = \{v_1, ..., v_{\rm N}\}$ are embeddings of $N$ corresponding image patches. Text Encoder extracts a sequence of text embeddings of $T$ as $E_t(T)=\{t_{\rm cls}, t_{\rm tok} \}$, where $t_{\rm cls}$ is the embedding of \texttt{[CLS]} token, and $t_{\rm tok} = \{t_1, ..., t_{\rm M}\}$ are embeddings of $M$ text tokens. 

\noindent \textbf{Manipulation-Aware Contrastive Learning with Global View.}
To help two uni-modal encoders better exploit the semantic correlation of images and texts, we align image and text embeddings through cross-modal contrastive learning. Nevertheless, some subtle multi-modal manipulations cause minor semantic inconsistency between two modalities, which are hardly unveiled by normal contrastive learning. To emphasize the semantic inconsistency caused by manipulations, we propose manipulation-aware contrastive learning on image and text embeddings. Different from normal cross-modal contrastive learning pulling embeddings of original image-text pairs close while only pushing those of unmatched pairs apart, manipulation-aware contrastive learning pushes away embeddings of manipulated pairs as well so that semantic inconsistency produced by them can be further emphasized. Following InfoNCE loss~\cite{oord2018representation}, we formulate image-to-text contrastive loss by:
\begin{equation}
\footnotesize
\begin{split}
\mathcal{L}_{v2t}(I, T^{+}, T^{-}) = -\mathbb{E}_{p(I,T)}\left[ {\rm log} \frac{{\rm exp}(S(I, T^{+})/\tau)}{\sum\nolimits_{k=1}^{K} {\rm exp} (S(I, T^{-}_{k})/\tau)} \right]
\end{split}
\end{equation}
where $\tau$ is a temperature hyper-parameter, $T^{-} = \{ T^{-}_1, ..., T^{-}_K\}$ is a set of negative text samples that are not matched to $I$ \textit{as well as that belong to manipulated image-text pairs}. Since \texttt{[CLS]} token serves as the semantic representation of the \textit{whole} image and text, we perform the cross-modal alignment based on them from the \textit{global} view as illustrated in Fig.~\ref{fig:MAC} (a). Specifically, we use two projection heads $h_v$ and $\hat{h}_t$ to map \texttt{[CLS]} tokens of two modalities to a lower-dimensional (256) embedding space for similarity calculation: $S(I, T)=h_v(v_{\rm cls})^{\rm T}\hat{h}_t(\hat{t}_{\rm cls})$. Inspired by MoCo~\cite{he2020momentum}, we learn momentum uni-modal encoders $\hat{E}_v$, $\hat{E}_t$ (an exponential-moving-average version) and momentum projection heads for two modalities respectively. Two queues are used to store the most recent $K$ image-text pair embeddings. Here $\hat{t}_{\rm cls}$ are $\texttt{[CLS]}$ tokens from text momentum encoders and $\hat{h}_t(\hat{t}_{\rm cls})$ means projected text embeddings from text momentum projection head. Similarly, text-to-image contrastive loss is as follows:
\begin{equation}
\footnotesize
\begin{split}
\mathcal{L}_{t2v}(T, I^{+}, I^{-}) = -\mathbb{E}_{p(I,T)}\left[ {\rm log} \frac{{\rm exp}(S(T, I^{+})/\tau)}{\sum\nolimits_{k=1}^{K} {\rm exp} (S(T, I^{-}_{k})/\tau)} \right]
\end{split}
\end{equation}
where $I^{-} = \{ I^{-}_1, ..., I^{-}_K\}$ is a queue of $K$ recent negative image samples that are not matched to $T$ \textit{as well as that belong to manipulated image-text pairs}. $S(T, I)=h_t(t_{\rm cls})^{\rm T}\hat{h}_v(\hat{v}_{\rm cls})$. Inspired by~\cite{yang2022vision}, to maintain reasonable semantic relation within every single modality, we further carry out intra-modal contrastive learning in both modalities. We incorporate all the losses to form Manipulation-Aware Contrastive Loss as follows:
\begin{equation}
\footnotesize
\begin{split}
\mathcal{L}_{MAC-G} = & \frac{1}{4} [\mathcal{L}_{v2t}(I, T^{+}, T^{-}) + \mathcal{L}_{t2v}(T, I^{+}, I^{-}) + \\
& \mathcal{L}_{v2v}(I, I^{+}, I^{-})+\mathcal{L}_{t2t}(T, T^{+}, T^{-}) ]
\end{split}
\end{equation}

\noindent \textcolor{black}{\noindent \textbf{Manipulation-Aware Contrastive Learning with Local View.}}
\textcolor{black}{Performing contrastive learning merely with a global view based on \texttt{[CLS]} token is likely to bring two drawbacks: \textbf{1)} Since manipulations in some image-text pairs are minor, a large portion of image regions and text tokens are not manipulated in these multi-modal media. Consequently, they should be more cross-modal semantic consistent than manipulated parts. For instance, as depicted in Fig.~\ref{fig:framework}, the person on the left corresponds to the name `Gordon Brown' in the text. Besides, surroundings in the image also imply they are in Paris, which matches `in Paris' mentioned in the text. In these cases, embeddings of such image regions and text tokens should not be pushed apart even though they are located in the manipulated pairs. Nevertheless, Manipulation-Aware Contrastive Learning with Global View only pushes the global semantic representations between two modalities away once encountering manipulated pairs. As such, these local image regions and text tokens without manipulation may serve as noise in the semantic alignment process; \textbf{2)} Since \texttt{[CLS]} token lacks image-specific inductive bias, \textit{e.g.,} spatial locality~\cite{dosovitskiy2020image}, the contrastive learning based on it ignores the local and structural information of the input.} 

\textcolor{black}{To alleviate these limitations, we further propose Manipulation-Aware Contrastive Learning with Local View, which performs more fine-grained semantic alignment between image-text pairs. It is well known that patch tokens ($v_{\rm pat}$ and $t_{\rm tok}$) from the transformer actually encode local features. Consequently, we could explore the embeddings of patch tokens to perform \textit{local} semantic alignment for each local image region or text token. In this regard, as illustrated in Fig.~\ref{fig:MAC} (b), we perform Manipulation-Aware Contrastive Learning between the embedding of \texttt{[CLS]} token and embeddings of patch tokens of two modalities. For an input image $I$, we consider the patch embeddings from paired original text $T_l^+=\hat{t}_{\rm tok}=\{\hat{t}_{i}\}_{i=1}^M$ as positive samples, while patch embeddings from unmatched texts as well as manipulated texts $T_l^-$ are used to build up negative samples. For an input text $T$, positive samples are set as patch embeddings from paired original image $I_l^+=\hat{v}_{\rm pat} =\{\hat{v}_{j}\}_{j=1}^N$, while negative samples are patch embeddings from unmatched images as well as manipulated images $I_l^-$. We can thus formulate Manipulation-Aware Contrastive Learning with Local View as follows,}
\textcolor{black}{\begin{equation}
\footnotesize
\begin{split}
\mathcal{L}_{MAC-L} = & \frac{1}{2} \left[ \sum\limits_{i=1}^{M} \mathcal{L}_{v2t}(I, T_{l(i)}^{+}, T_{l}^{-}) + \sum\limits_{j=1}^{N}\mathcal{L}_{t2v}(T, I_{l(j)}^{+}, I_{l}^{-}) \right]
\end{split}
\end{equation}}
\textcolor{black}{where $S(I, T_{l(i)}^{+})=h_v(v_{\rm cls})^{\rm T}\hat{h}_t(\hat{t}_{i})$, $S(T, I_{l(j)}^{+})=h_t(t_{\rm cls})^{\rm T}\hat{h}_v(\hat{v}_{j})$, $T_{l}^{-}$ and $I_{l}^{-}$ are in-batch negative text and image patch embeddings, respectively. Therefore, minimizing $\mathcal{L}_{MAC-L}$ facilitates the adaptive cross-modal semantic alignment between each local patch embedding and the global embedding of \texttt{[CLS]} token. This allows our model to explicitly weigh the importance of each local image region or text token in the semantic alignment process and thus contributes to more fine-grained cross-modal relation reasoning between two modalities.}

\noindent \textbf{Manipulated Image Bounding Box Grounding.}
As mentioned above, FS or FA swaps identities or edits attributes of faces in images. This alters their correlation to corresponding texts in terms of persons' names or emotions. Given this, we argue the manipulated image region could be located by finding local patches that have inconsistencies with text embeddings. In this regard, we perform cross-attention between image and text embeddings to obtain patch embeddings that contain image-text correlation. Attention function~\cite{vaswani2017attention} is performed on normalized query ($Q$), key ($K$), and value ($V$) features as:
\begin{equation}
\footnotesize
\begin{split}
\text{Attention}(Q, K, V) =\text{Softmax}(K^T Q / \sqrt{D})V
\end{split}
\end{equation}
Here we cross-attend the image embedding with text embedding, by treating $Q$ as image embedding, $K$ and $V$ as text embedding as follows:
\begin{equation}
\footnotesize
\begin{split}
U_v(I) & = \text{Attention}(E_v(I), E_t(T), E_t(T)) + E_v(I)
\end{split}
\end{equation}
where $U_v(I)=\{u_{\rm cls}, u_{\rm pat} \}$. $u_{\rm pat} = \{u_1, ..., u_{\rm N} \}$ are $N$ image patch embeddings interacted with text information. Unlike \texttt{[CLS]} token $u_{\rm cls}$, the patch tokens $u_{\rm pat}$ are generated with position encoding~\cite{vaswani2017attention}. This means they possess richer local spatial information and thus are more suitable for manipulated image bbox grounding. Based on this analysis, we propose \textit{Local Patch Attentional Aggregation (\textbf{LPAA})} to aggregate the spatial information in $u_{\rm pat}$ via an attentional mechanism. This aggregation is performed by cross-attending a $\texttt{[AGG]}$ token with $u_{\rm pat}$ as follows:
\begin{equation}
\footnotesize
\begin{split}
u_{\rm agg} = \textbf{LPAA}(u_{\rm pat})=\text{Attention}(\texttt{[AGG]}, u_{\rm pat}, u_{\rm pat})
\end{split}
\end{equation}
where $\texttt{[AGG]}$ token serves as the token adaptively aggregating spatial information of $u_{\rm pat}$. Different from previous work~\cite{zeng2021multi} directly using \texttt{[CLS]} token for bbox prediction, we perform the manipulated bbox grounding based on the attentionally aggregated embedding $u_{\rm agg}$. Specifically, we input $u_{\rm agg}$ into BBox Detector $D_v$ and calculate Image Manipulation Grounding Loss by combing normal $\ell_{1}$ loss and generalized Intersection over Union (IoU) loss~\cite{rezatofighi2019generalized} as follows:
\begin{equation}
\footnotesize
\begin{split}
\mathcal{L}_{IMG} = & \mathbb{E}_{(I,T)\sim P} [\| {\rm Sigmoid}(D_v(u_{\rm agg}))-y_{box}\| \\
& + \mathcal{L}_{\rm IoU}({\rm Sigmoid}(D_v(u_{\rm agg}))-y_{box}) ]
\end{split}
\end{equation}
\subsection{Deep Manipulation Reasoning}
Manipulated token grounding is a tougher task than manipulated bbox grounding as it requires deeper analysis and reasoning on the correlation between images and texts. For example, as illustrated in Fig.~\ref{fig:framework}, we are able to detect the manipulated tokens in $T$, \textit{i.e.,} `force' and `resign', only when we are aware of such negative words mismatching the positive emotion (\textit{i.e.,} smiling faces) in $I$. Besides, we need to summarize multi-modal information to detect fine-grained manipulation types and binary classes. This demands a comprehensive information summary at this stage. To this end, we propose deep manipulation reasoning.

\noindent \textbf{Manipulated Text Token Grounding.}
To model deeper multi-modal interaction, as depicted in Fig.~\ref{fig:framework}, we propose modality-aware cross-attention to further lead text embedding $E_t(T)$ to interact with image embedding $E_v(I)$ through multiple cross-attention layers in Multi-Modal Aggregator $F$. This generates aggregated multi-modal embedding $F(E_v(I), E_t(T))=\{m_{\rm cls}, m_{\rm tok}\}$. In particular, $m_{\rm tok} = \{m_1,..., m_{\rm M} \}$ represent the deeper aggregated embeddings corresponding to each token in $T$. At this stage, each token in $T$ has passed through multiple self-attention layers in $E_t$ and cross-attention layers in $F$. In this way, each token embedding in $m_{\rm tok}$ not only entirely explores the context information of text, but also fully interacts with image features, which fits manipulated text tokens grounding. Moreover, grounding manipulated tokens is equal to labeling each token as real or fake. This is similar to sequence tagging task in NLP. Notably, unlike existing sequence tagging task mainly studied in text modality, manipulated text token grounding here can be regarded as a novel \textit{multi-modal sequence tagging} since each token is interacted with two modality information. In this case, we use a Token Detector $D_t$ to predict the label of each token in $m_{\rm tok}$ and calculate cross-entropy loss as follows:
\begin{equation}
\footnotesize
\begin{split}
\mathcal{L}_{tok} = \mathbb{E}_{(I,T)\sim P}{\rm \textbf{H}}(D_t(m_{\rm tok}), y_{\rm tok})
\end{split}
\end{equation}
where ${\rm \textbf{H}(\cdot)}$ is the cross-entropy function. As mentioned, news on the web is usually noisy with texts unrelated to paired images~\cite{li2021align}. To alleviate over-fitting to noisy texts, as shown in Fig.~\ref{fig:framework}, we further learn momentum versions for Multi-Modal Aggregator and Token Detector, respectively, denoted as $\hat{F}$ and  $\hat{D}_t$. We can obtain the multi-modal embedding from momentum modules as $\hat{F}(\hat{E}_v(I), \hat{E}_t(T))=\{\hat{m}_{\rm cls}, \hat{m}_{\rm tok}\}$. Based on this, momentum Token Detector generates soft pseudo-labels to modulate the original token prediction, by calculating the KL-Divergence as follows:
\begin{equation}
\footnotesize
\begin{split}
\mathcal{L}_{tok}^{mom} = \mathbb{E}_{(I,T)\sim P}{\rm KL} \left[D_t(m_{\rm tok})\| \hat{D}_t(\hat{m}_{\rm tok}) \right]
\end{split}
\end{equation}
The final Text Manipulation Grounding Loss is a weighted combination as follows:
\begin{equation}
\footnotesize
\begin{split}
\mathcal{L}_{TMG} = (1- \alpha)\mathcal{L}_{tok}+\alpha \mathcal{L}_{tok}^{mom}
\end{split}
\end{equation}

\noindent \textbf{Fine-Grained Manipulation Type Detection and Binary Classification.}
Unlike current forgery detection works mainly performing real/fake binary classification, we expect our model to provide more interpretation for manipulation detection. As mentioned in Sec.~\ref{sec:Dataset_Manipulation}, two image and two text manipulation approaches are introduced in \textbf{DGM$^4$} dataset. Given this, we aim to further detect four fine-grained manipulation types. As different manipulation types could appear in one image-text pair simultaneously, we treat this task as a specific \textit{multi-modal multi-label classification}. Since \texttt{[CLS]} token $m_{\rm cls}$ aggregates multi-modal information after modality-aware cross-attention, it can be utilized as a comprehensive summary of manipulation characteristics. We thus concatenate a Multi-Label Classifier $C_m$ on top of it to calculate Multi-Label Classification Loss:
\begin{equation}
\footnotesize
\begin{split}
\mathcal{L}_{MLC} = \mathbb{E}_{(I,T)\sim P}{\rm \textbf{H}}(C_m(m_{\rm cls}), y_{\rm mul})
\end{split}
\end{equation}
Naturally, we also conduct a normal binary classification based on $m_{\rm cls}$ as follows:
\begin{equation}
\footnotesize
\begin{split}
\mathcal{L}_{BIC} = \mathbb{E}_{(I,T)\sim P}{\rm \textbf{H}}(C_b(m_{\rm cls}), y_{\rm bin})
\end{split}
\end{equation}

\begin{table*}[t]
\scriptsize
\centering
\caption{Comparison of multi-modal learning methods for DGM$^{4}$.}
\begin{tabular}{c|ccc|ccc|ccc|ccc}
\toprule[1pt]
Categories & \multicolumn{3}{c|}{Binary Cls}    & \multicolumn{3}{c|}{Multi-Label Cls}  & \multicolumn{3}{c|}{Image Grounding} & \multicolumn{3}{c}{Text Grounding} \\ \hline
Methods              & AUC            & EER            & ACC            & mAP            & CF1            & OF1            & IoUmean       & IoU50        & IoU75        & Precision      & Recall         & F1             \\ \hline
CLIP~\cite{radford2021learning}                 & 83.22          & 24.61          & 76.40          & 66.00          & 59.52          & 62.31          & 49.51           & 50.03          & 38.79          & 58.12          & 22.11          & 32.03          \\
ViLT~\cite{kim2021vilt}                 & 85.16          & 22.88          & 78.38          & 72.37          & 66.14          & 66.00          & 59.32           & 65.18          & 48.10          & 66.48          & 49.88          & 57.00          \\
\rowcolor[HTML]{E3DCDC} 
\textbf{Ours}        & 93.19 & 14.10 & 86.39 & 86.22 & 79.37 & 80.37 & 76.45  & 83.75 & \textbf{76.06} & \textbf{75.01} & 68.02 & 71.35 \\ 
\rowcolor[HTML]{E3DCDC} 
\textbf{\textcolor{black}{Ours++}}        & \textcolor{black}{\textbf{93.33}} & \textbf{\textcolor{black}{14.06}} & \textcolor{black}{\textbf{86.66}} & \textcolor{black}{\textbf{86.41}} & \textcolor{black}{\textbf{79.73}} & \textcolor{black}{\textbf{80.71}} & \textcolor{black}{\textbf{76.46}}  & \textcolor{black}{\textbf{83.77}} & \textcolor{black}{76.03} & \textcolor{black}{73.05} & \textcolor{black}{\textbf{72.14}} & \textcolor{black}{\textbf{72.59}} \\ \bottomrule[1pt]
\end{tabular}
\label{tab:multi-modal} 
\end{table*}

\begin{table}[t]
\scriptsize
\centering
\caption{Comparison of deepfake detection methods for DGM$^{4}$.}
\begin{tabular}{c|ccc|ccc}
\toprule[1pt]
Categories     & \multicolumn{3}{c|}{Binary Cls}               & \multicolumn{3}{c}{Image Grounding}         \\ \hline
Methods        & AUC            & EER            & ACC            & IoUmean      & IoU50        & IoU75        \\ \hline
TS~\cite{luo2021generalizing} & 91.80          & 17.11          & 82.89          & 72.85          & 79.12          & 74.06          \\
MAT~\cite{zhao2021multi}            & 91.31          & 17.65          & 82.36          & 72.88          & 78.98          & 74.70          \\
\rowcolor[HTML]{E3DCDC} 
\textbf{Ours}  & 94.40 & 13.18 & 86.80 & 75.69 & 82.93 & 75.65\\ 
\rowcolor[HTML]{E3DCDC} 
\textbf{\textcolor{black}{Ours++}}  & \textcolor{black}{\textbf{94.69}} & \textcolor{black}{\textbf{13.04}} & \textcolor{black}{\textbf{86.82}} & \textcolor{black}{\textbf{75.96}} & \textcolor{black}{\textbf{83.32}} & \textcolor{black}{\textbf{75.80}}\\\bottomrule[1pt]
\end{tabular}
\label{tab:deepfake} 
\end{table}

\begin{table}[t]
\scriptsize
\centering
\caption{Comparison of sequence tagging methods for DGM$^{4}$.}
\begin{tabular}{c|ccc|ccc}
\toprule[1pt]
Categories    & \multicolumn{3}{c|}{Binary Cls}               & \multicolumn{3}{c}{Text Grounding}         \\ \hline
Methods       & AUC            & EER            & ACC            & Precision      & Recall         & F1             \\ \hline
BERT~\cite{devlin2019bert}          & 80.82          & 28.02          & 68.98          & 41.39          & 63.85          & 50.23          \\
LUKE~\cite{yamada2020luke}          & 81.39          & 27.88          & 76.18          & 50.52          & 37.93          & 43.33          \\
\rowcolor[HTML]{E3DCDC} 
\textbf{Ours} & 93.44 & 13.83 & 87.39 & 70.90 & \textbf{73.30} & 72.08 \\ 
\rowcolor[HTML]{E3DCDC} 
\textbf{\textcolor{black}{Ours++}} & \textcolor{black}{\textbf{93.49}} & \textcolor{black}{\textbf{13.58}} & \textcolor{black}{\textbf{87.81}} & \textcolor{black}{\textbf{72.70}} & \textcolor{black}{72.57} & \textcolor{black}{\textbf{72.64}} \\ \bottomrule[1pt]
\end{tabular}
\label{tab:seqtag} 
\end{table}

\begin{table}[t]
\scriptsize
\centering
\caption{Ablation study of image modality.}
\begin{tabular}{c|ccc|ccc}
\toprule[1pt]
Categories    & \multicolumn{3}{c|}{Binary Cls}               & \multicolumn{3}{c}{Image Grounding}              \\ \hline
Methods       & AUC            & EER            & ACC            & IoUmean        & IoU50          & IoU75          \\ \hline
Ours-Image   & 93.96          & 13.83          & 86.13          & 75.58         & 82.44          & 75.80          \\
\textbf{Ours}  & 94.40 & 13.18 & 86.80 & 75.69 & 82.93 & 75.65\\ 
\textcolor{black}{\textbf{Ours++}}  & \textcolor{black}{\textbf{94.69}} & \textcolor{black}{\textbf{13.04}} & \textcolor{black}{\textbf{86.82}} & \textcolor{black}{\textbf{75.96}} & \textcolor{black}{\textbf{83.32}} & \textcolor{black}{\textbf{75.80}}\\
\bottomrule[1pt]
\end{tabular}
\label{tab:abl_img} 
\end{table}

\begin{table}[t]
\scriptsize
\centering
\caption{Ablation study of text modality.}
\begin{tabular}{c|ccc|ccc}
\toprule[1pt]
Categories    & \multicolumn{3}{c|}{Binary Cls}               & \multicolumn{3}{c}{Text Grounding}         \\ \hline
Methods       & AUC            & EER            & ACC            & Precision      & Recall         & F1             \\ \hline
Ours-Text    & 75.67          & 32.46          & 72.17          & 42.99          & 33.68          & 37.77          \\
\textbf{Ours} & 93.44 & 13.83 & 87.39 & 70.90 & \textbf{73.30} & 72.08 \\ 
\textcolor{black}{\textbf{Ours++}} & \textcolor{black}{\textbf{93.49}} & \textcolor{black}{\textbf{13.58}} & \textcolor{black}{\textbf{87.81}} & \textcolor{black}{\textbf{72.70}} & \textcolor{black}{72.57} & \textcolor{black}{\textbf{72.64}} \\ \bottomrule[1pt]
\end{tabular}
\label{tab:abl_txt} 
\end{table}

\section{Experiments}
\label{sec:experiments}
\subsection{Experimental Setup}
\label{sec:Experimental Setup}
\textcolor{black}{\noindent \textbf{Implementation Details.} All of our experiments are performed on 8 NVIDIA V100 GPUs with PyTorch framework~\cite{paszke2017automatic}. Image Encoder is implemented by ViT-B/16~\cite{dosovitskiy2020image} with 12 layers. Both Text Encoder and Multi-Modal Aggregator are built based on a 6-layer transformer initialized by the first 6 layers and the last 6 layers of BERT$_{base}$~\cite{devlin2019bert}, respectively. Binary Classifier, Multi-Label Classifier, BBox Detector, and Token Detector are set up to two Multi-Layer Perception (MLP) layers with output dimensions as 2, 4, 4, and 2. We set queue size $K = 65, 536$. AdamW~\cite{loshchilov2017decoupled} optimizer is adopted with a weight decay of 0.02. The learning rate is warmed-up to $1e^{-4}$ in the first 1000 steps, and decayed to $1e^{-5}$ following a cosine schedule.}

\noindent \textcolor{black}{\noindent \textbf{Evaluation Metrics.}
To evaluate the proposed new research problem DGM$^{4}$ comprehensively, we set up rigorous evaluation protocols and metrics for all the manipulation detection and grounding tasks.
\begin{itemize}[leftmargin=*]
\item \textbf{Binary classification:} Following current deepfake methods~\cite{luo2021generalizing,zhao2021multi}, we adopt Accuracy (ACC), Area Under the Receiver Operating Characteristic Curve (AUC), and Equal Error Rate (EER) for evaluation of binary classification.
\item \textbf{Multi-label classification:} Like existing multi-label classification methods~\cite{ben2020asymmetric,durand2019learning}, we use mean Average Precision (MAP), average per-class F1 (CF1), and average overall F1 (OF1) for evaluating the detection of fine-grained manipulation types.
\item \textbf{Manipulated image bounding box grounding:} To examine the performance of predicted manipulated bbox, we calculate the mean of Intersection over Union (IoUmean) between ground-truth and predicted coordinates of all testing samples. Moreover, we set two thresholds (0.5, 0.75) of IoU and calculate the average accuracy (correct grounding if IoU is above the threshold and versa vice), which are denoted as IoU50 and IoU75.
\item \textbf{Manipulated text token grounding:} Considering the class imbalance scenario that manipulated tokens are much fewer than original tokens, we adopt Precision, Recall, F1 Score as metrics. This contributes to a more fair and reasonable evaluation for manipulated text token grounding.
\end{itemize}}

\begin{table*}[t]
\scriptsize
\centering
\caption{Ablation study of losses in the proposed method.}
\begin{tabular}{p{0.3cm}<{\centering}p{0.3cm}<{\centering}p{0.88cm}<{\centering}p{0.3cm}<{\centering}p{0.3cm}<{\centering}p{0.88cm}<{\centering}|ccc|ccc|ccc|ccc}
\toprule[1pt]
\multicolumn{6}{c|}{Losses}                                                                                                                                   & \multicolumn{3}{c|}{Binary Cls}               & \multicolumn{3}{c|}{Multi-Label Cls}             & \multicolumn{3}{c|}{Image Grounding}            & \multicolumn{3}{c}{Text Grounding}               \\ \hline
BIC                           & MLC                           & MAC-G                           & IMG                           & TMG & MAC-L                          & AUC            & EER            & ACC            & mAP            & CF1            & OF1            & IoUmean      & IoU50        & IoU75        & Precision      & Recall         & F1             \\ \hline
\CheckmarkBold & &    &    &    &    & 91.04          & 16.91          & 83.81          & 20.79          & 33.84         & 33.48          & 4.81          & 0.33           & 0.00           & 15.95          & \textbf{78.70}          & 26.53          \\
\CheckmarkBold &    & \CheckmarkBold & \CheckmarkBold & \CheckmarkBold &  & 91.74          & 16.08          & 84.39          & 27.22          & 30.81          & 27.62          & 74.05         & 81.34          & 72.59          & 74.30          & 66.84          & 70.37          \\
\CheckmarkBold & \CheckmarkBold &    & \CheckmarkBold & \CheckmarkBold & & 92.77          & 14.53          & 86.01          & 85.52          & 79.09          & 79.86          & 75.98         & 83.37          & 75.25          & \textbf{77.82} & 61.83          & 68.91          \\
\CheckmarkBold & \CheckmarkBold & \CheckmarkBold &    & \CheckmarkBold & & 93.21          & 14.30           & 86.28          & 86.29 & 79.37 & 80.32 & 4.69          & 0.17           & 0.00         & 75.72          & 67.44 & 71.34 \\
\CheckmarkBold & \CheckmarkBold & \CheckmarkBold & \CheckmarkBold & &    & 92.99           & 14.62          & 86.15          & 86.06          & 79.06         & 79.93          & \textbf{76.51} & 83.73 & 76.05          & 13.93          & 58.87          & 22.53           \\
\CheckmarkBold & \CheckmarkBold & \CheckmarkBold & \CheckmarkBold & \CheckmarkBold & & 93.19 & 14.10 & 86.39 & 86.22          & 79.37          & 80.37          & 76.45         & 83.75          & \textbf{76.06} & 75.01          & 68.02          & 71.35          \\ 
\CheckmarkBold & \CheckmarkBold & \CheckmarkBold & \CheckmarkBold & \CheckmarkBold & \CheckmarkBold& \textcolor{black}{\textbf{93.33}} & \textcolor{black}{\textbf{14.06}} & \textcolor{black}{\textbf{86.66}} & \textcolor{black}{\textbf{86.41}}          & \textcolor{black}{\textbf{79.73}}          & \textcolor{black}{\textbf{80.71}}          & \textcolor{black}{76.46}         & \textcolor{black}{\textbf{83.77}}         & \textcolor{black}{76.03} & \textcolor{black}{73.05}         & \textcolor{black}{72.14}          & \textcolor{black}{\textbf{72.59}}          \\\bottomrule[1pt]
\end{tabular}
\label{tab:abl_loss} 
\end{table*}

\begin{figure}[t]
    \centering
     \begin{minipage}[t]{0.55\linewidth}
     \captionsetup{width=0.95\textwidth}
        \centering
        \includegraphics[height=90pt]{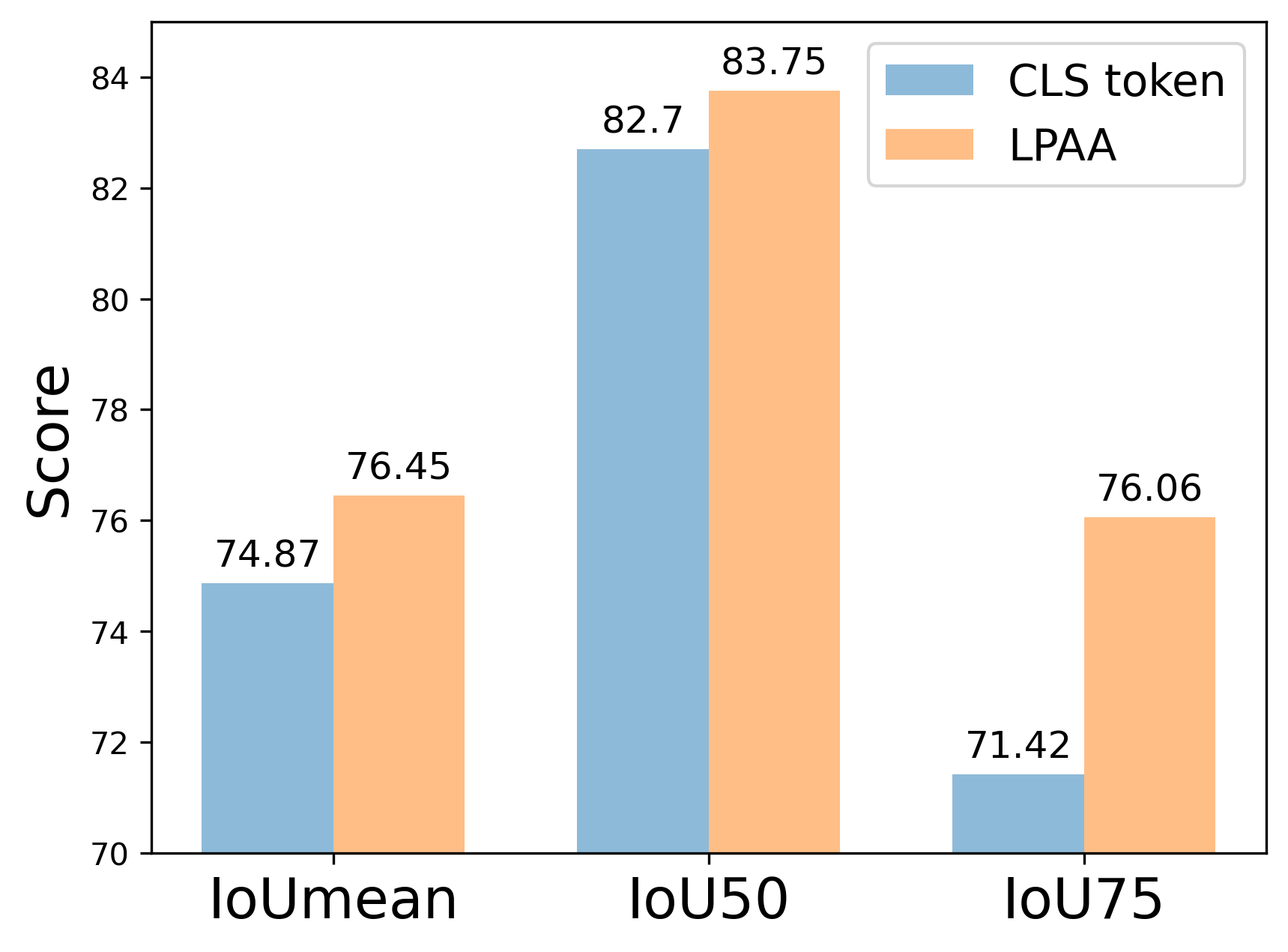}
        \caption{Efficacy of local patch attentional aggregation (LPAA).}
        \label{fig:abl_lpaa}
    \end{minipage}
    \begin{minipage}[t]{0.40\linewidth}
        \centering
        \includegraphics[height=90pt]{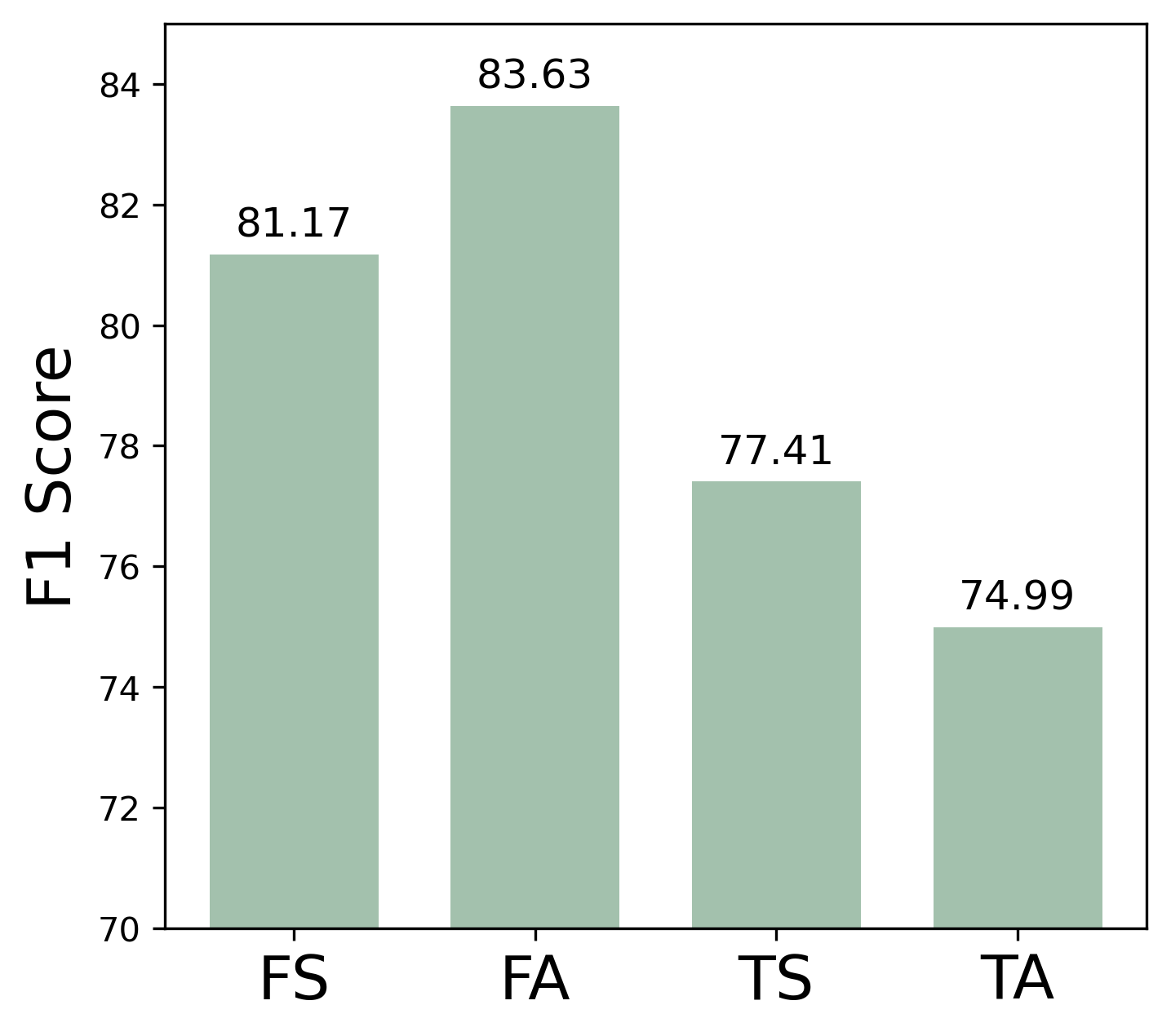}
        \caption{Performance of each manipulation type.}
        \label{fig:multicls}
    \end{minipage}
\end{figure}

\begin{figure*}[t]
    \centering
    \subfloat[Ours-Image]{\includegraphics[width = 0.48\textwidth]{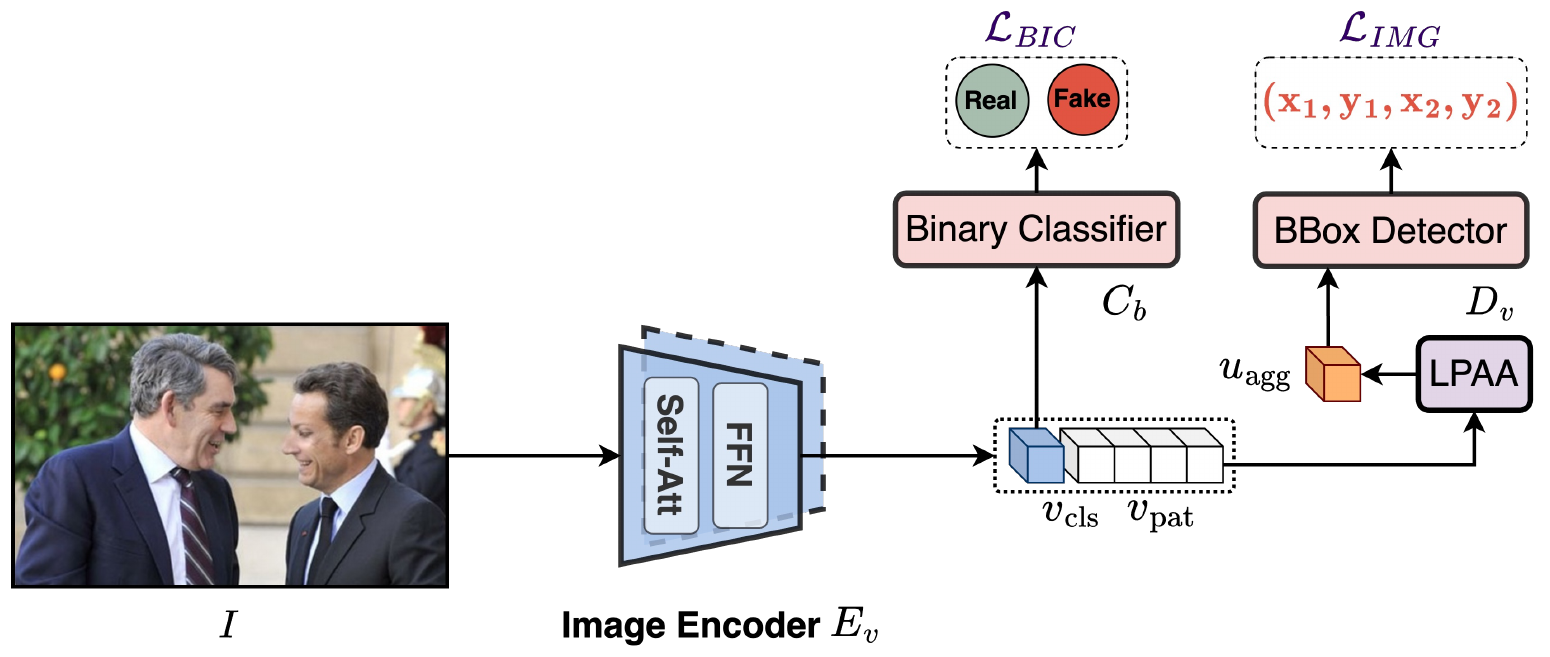}}
    \label{fig:framework_img_abl}
    \hfill
    \subfloat[Ours-Text]{\includegraphics[width = 0.48\textwidth]{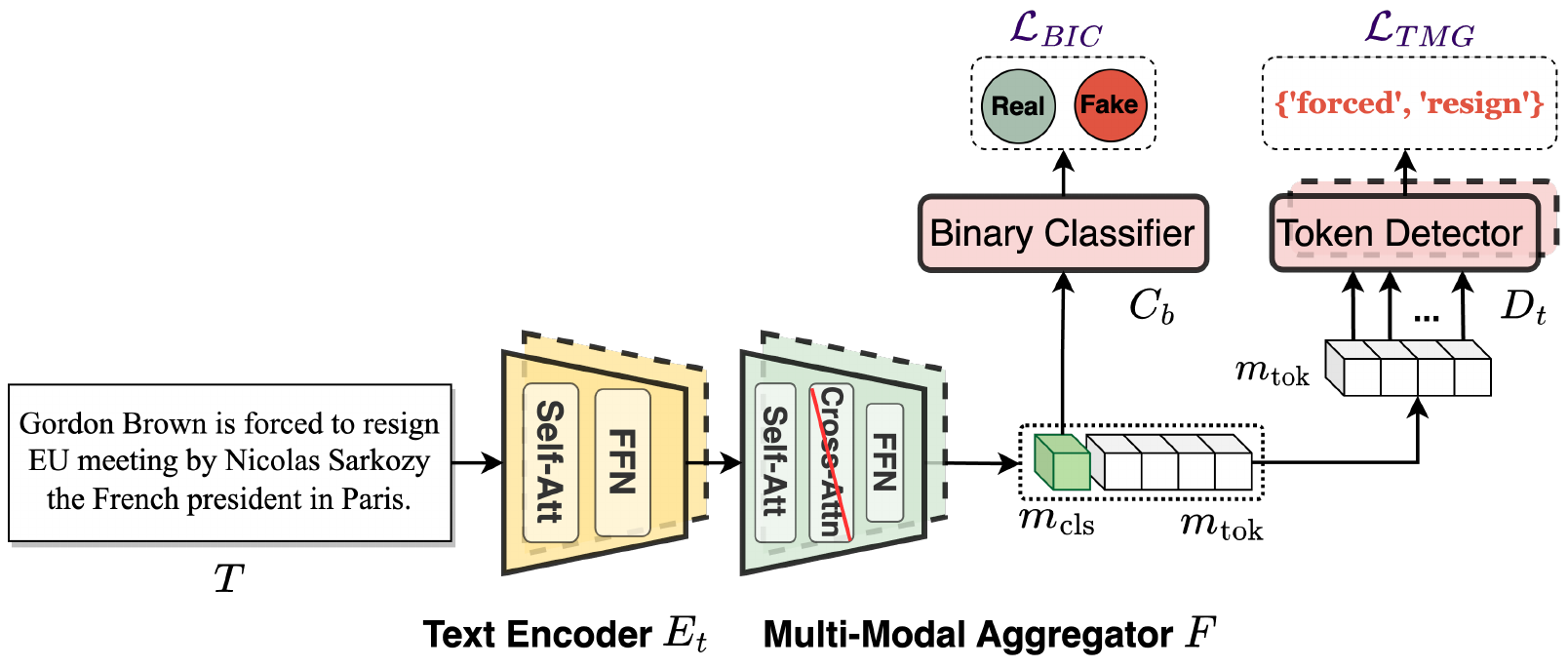}}
    \label{fig:framework_text_abl}
    \caption{\textcolor{black}{Frameworks of ablation studies in (a) image modality and (b) text modality. We remove the input and network components of one modality when we perform an ablation study on the other.}}
    \label{fig:ablation_study}
\end{figure*}

\begin{table*}[t]
\scriptsize
\centering
\caption{\textcolor{black}{Comparison between manipulation-aware contrastive learning and normal contrastive learning.}}
\textcolor{black}{\begin{tabular}{c|ccc|ccc|ccc|ccc}
\toprule[1pt]
Categories                     & \multicolumn{3}{c|}{Binary Cls}                  & \multicolumn{3}{c|}{Multi-Label Cls}             & \multicolumn{3}{c|}{Image Grounding}             & \multicolumn{3}{c}{Text Grounding}               \\ \hline
Methods                        & AUC            & EER            & ACC            & mAP            & CF1            & OF1            & IoU\_mean      & IoU\_50        & IoU\_75        & Precision      & Recall         & F1             \\ \hline
Normal CL & 92.95          & 14.34          & 86.24          & 85.78          & 79.29          & 80.29          & 76.41          & 83.67          & 76.01          & 72.35          & 71.58 & 71.97 \\
\rowcolor[HTML]{E3DCDC} 
\textbf{Manipulation-Aware CL-G}                  & 93.19 & 14.10 & 86.39 & 86.22 & 79.37 & 80.37 & 76.45 & 83.75 & \textbf{76.06} & \textbf{75.01} & 68.02          & 71.35          \\ 
\rowcolor[HTML]{E3DCDC} 
\textbf{\textcolor{black}{Manipulation-Aware CL-G+L}}                 & \textcolor{black}{\textbf{93.33}} & \textbf{\textcolor{black}{14.06}} & \textcolor{black}{\textbf{86.66}} & \textcolor{black}{\textbf{86.41}} & \textcolor{black}{\textbf{79.73}} & \textcolor{black}{\textbf{80.71}} & \textcolor{black}{\textbf{76.46}}  & \textcolor{black}{\textbf{83.77}} & \textcolor{black}{76.03} & \textcolor{black}{73.05} & \textcolor{black}{\textbf{72.14}} & \textcolor{black}{\textbf{72.59}}         \\ \bottomrule[1pt]
\end{tabular}}
\label{tab:abl_manipulation_aware}
\end{table*}

\begin{table}[t]
\scriptsize
\centering
\caption{\textcolor{black}{Ablation study of momentum models.}}
\textcolor{black}{\begin{tabular}{c|ccc|ccc}
\toprule[1pt]
Categories    & \multicolumn{3}{c|}{Binary Cls}                  & \multicolumn{3}{c}{Text Grounding}               \\ \hline
Methods       & AUC            & EER            & ACC            & Precision      & Recall         & F1             \\ \hline
Ours w/o Mom  & 92.96          & 14.47          & 86.13          & \textbf{76.13} & 66.99          & 71.27          \\
\textbf{Ours} & \textbf{93.19} & \textbf{14.10} & \textbf{86.39} & 75.01          & \textbf{68.02} & \textbf{71.35} \\ \bottomrule[1pt]
\end{tabular}}
\label{tab:abl_mom}
\end{table}

\subsection{Benchmark for \texorpdfstring{DGM$^{4}$}{Lg}}
\noindent \textbf{Comparison with multi-modal learning methods.} 
We adapt two SOTA multi-modal learning methods to \textbf{DGM$^{4}$} setting for comparison. Specifically, CLIP~\cite{radford2021learning} is one of the most popular \textit{dual-stream} approaches where two modalities are not concatenated at the input level. For adaptation, we make outputs of two streams interact with each other through cross-attention layers. Detection and grounding heads are further integrated on top of them. In addition, ViLT~\cite{kim2021vilt} is a representative \textit{single-stream} approach where cross-modal interaction layers are operated on a concatenation of image and text inputs. We also adapt it by concatenating detection and grounding heads on corresponding outputs of the model. We tabulate comparison results in Table~\ref{tab:multi-modal}. The results show that the \textbf{HAMMER} significantly outperforms both baselines in terms of all evaluation metrics. This demonstrates that hierarchical manipulation reasoning is more able to accurately and comprehensively model the correlation between images and texts and capture semantically inconsistency caused by manipulation, contributing to better manipulation detection and grounding. \textcolor{black}{Moreover, as illustrated in Table~\ref{tab:multi-modal}, \textbf{HAMMER++} further improves the performance in most of evaluation metrics. This implies through adaptively aligning each local patch embedding with global class token, the supplemented Manipulation-Aware Contrastive Learning with Local View yields more fine-grained semantic alignment, promoting the manipulation detection and grounding tasks.}

\noindent \textbf{Comparison with deepfake detection and sequence tagging methods.}
We compare our method with competitive uni-modal methods in two single-modal forgery data splits, respectively. For a fair comparison, in addition to the original ground-truth regarding binary classification, we further integrate manipulation grounding heads into uni-modal models with corresponding annotations of grounding. For image modality, we tabulate the comparison with two SOTA deepfake detection methods in Table~\ref{tab:deepfake}. For text modality, we compare two widely-used sequence tagging methods in NLP to ground manipulated tokens along with binary classification. We report the comparison results in Table~\ref{tab:seqtag}. Tables~\ref{tab:deepfake} and~\ref{tab:seqtag} show that \textbf{HAMMER} performs better than uni-modal methods for single-modal forgery detection by a large margin, indicating our method trained with multi-modal media also achieves promising manipulation detection and grounding performance in each single modality. \textcolor{black}{In addition, \textbf{HAMMER++} further substantially exceeds \textbf{HAMMER} in both uni-modal evaluation scenarios. This suggests Manipulation-Aware Contrastive Learning with Local View facilities manipulation detection and grounding performance in each modality as well.}
\begin{figure}[t]
    \centering
    \subfloat[GT: \textcolor{Red}{Fake-FS}, Pred: \textcolor{RoyalBlue}{Fake-FS}]{\includegraphics[width = 0.24\textwidth]{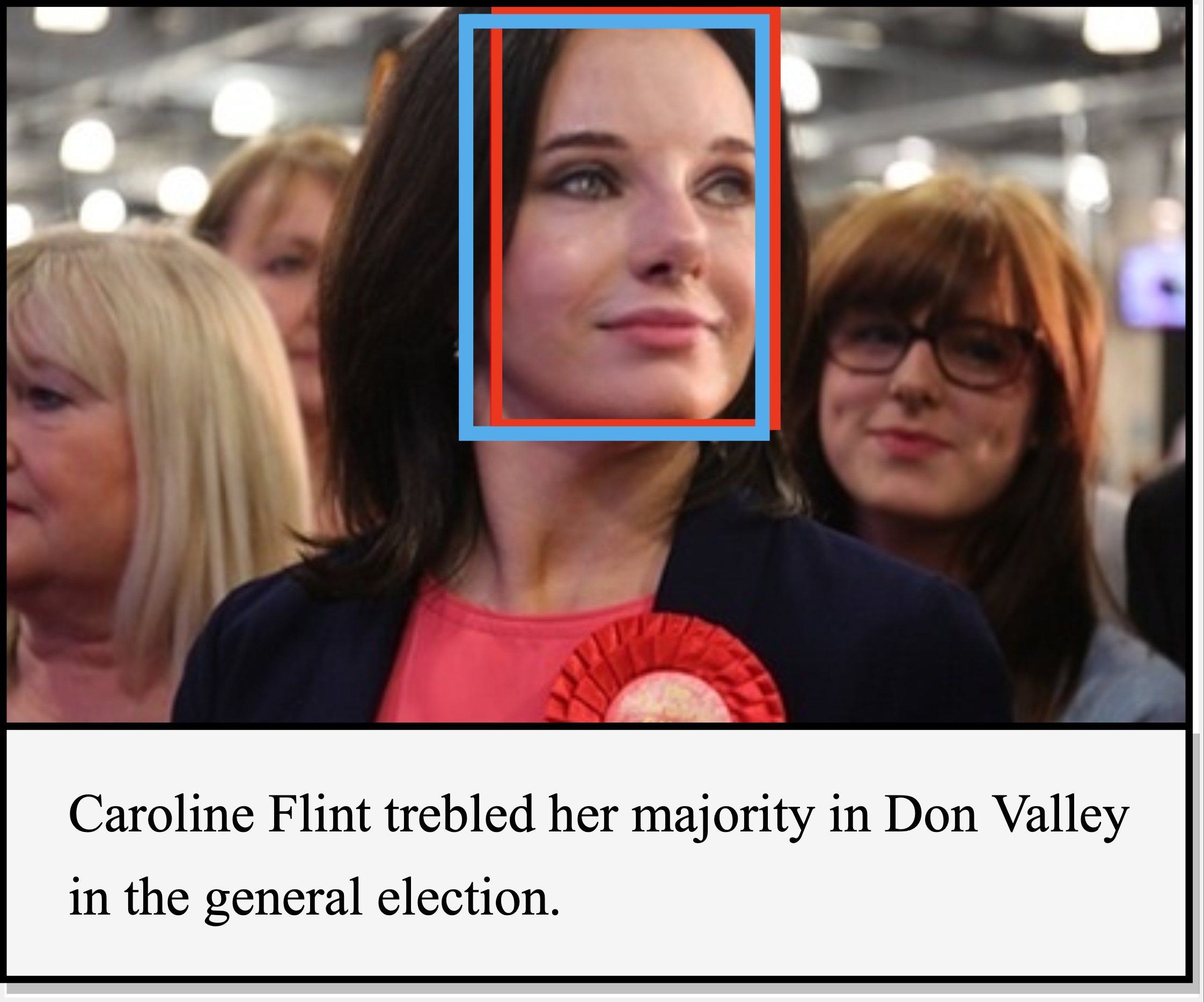}}
    \label{fig:vis_FS}
    \hfill
    \subfloat[GT: \textcolor{Red}{Fake-FA}, Pred: \textcolor{RoyalBlue}{Fake-FA}]{\includegraphics[width = 0.24\textwidth]{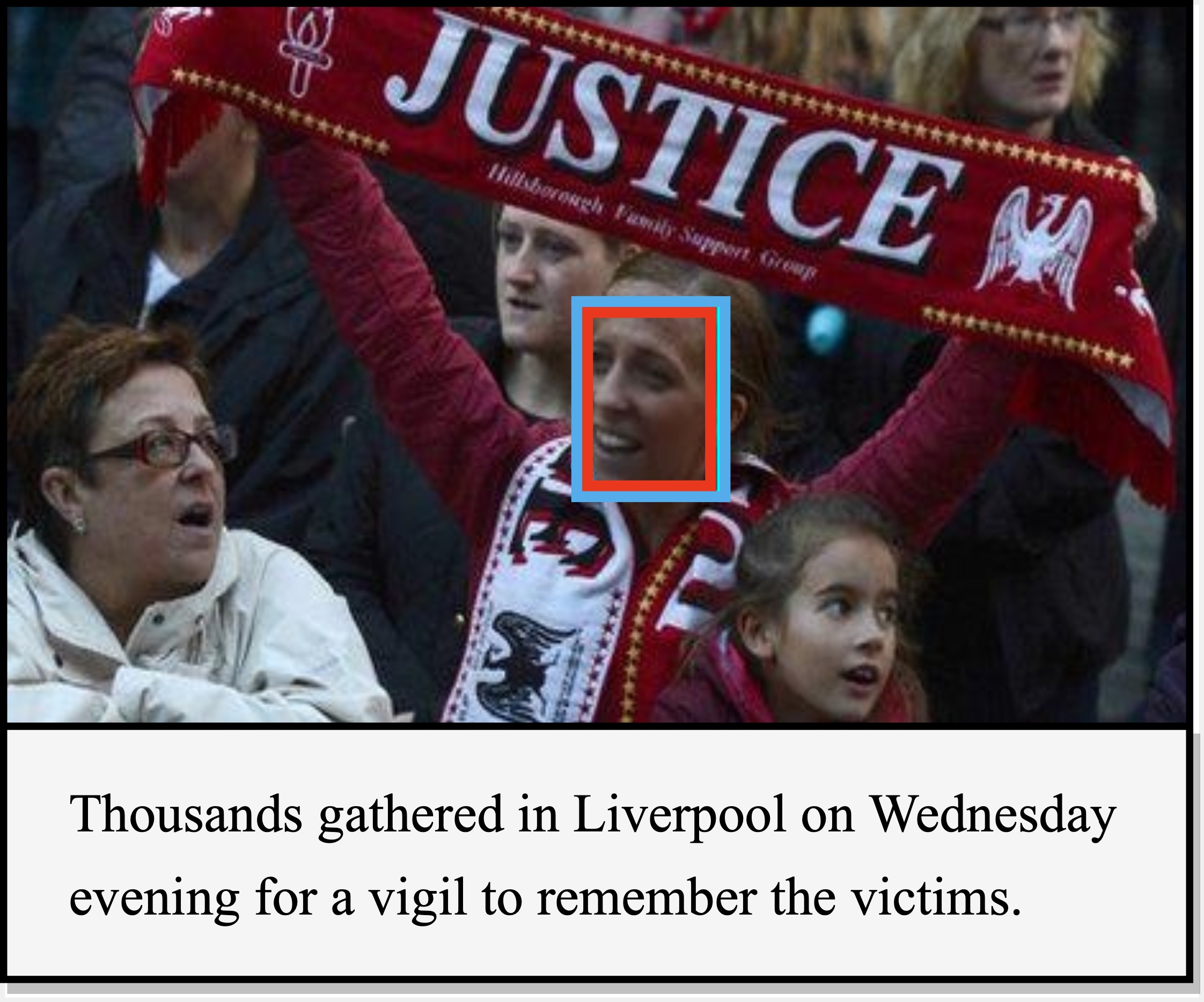}}
    \label{fig:vis_FA}
    \hfill
    \subfloat[GT: \textcolor{Red}{Fake-TS}, Pred: \textcolor{RoyalBlue}{Fake-TS}]{\includegraphics[width = 0.24\textwidth]{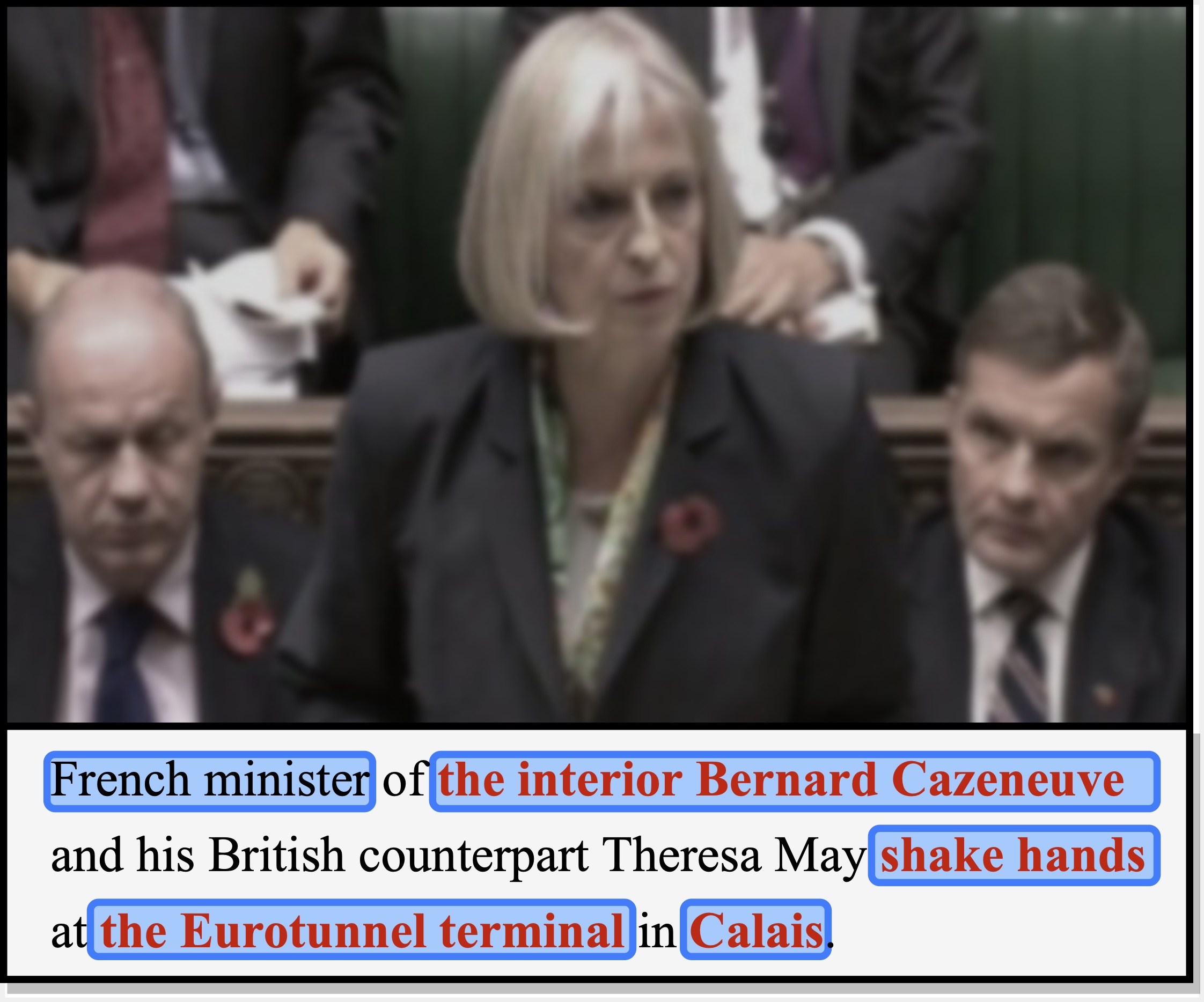}}
    \label{fig:vis_TS}
    \hfill
    \subfloat[GT: \textcolor{Red}{Fake-TA}, Pred: \textcolor{RoyalBlue}{Fake-TA}]{\includegraphics[width = 0.24\textwidth]{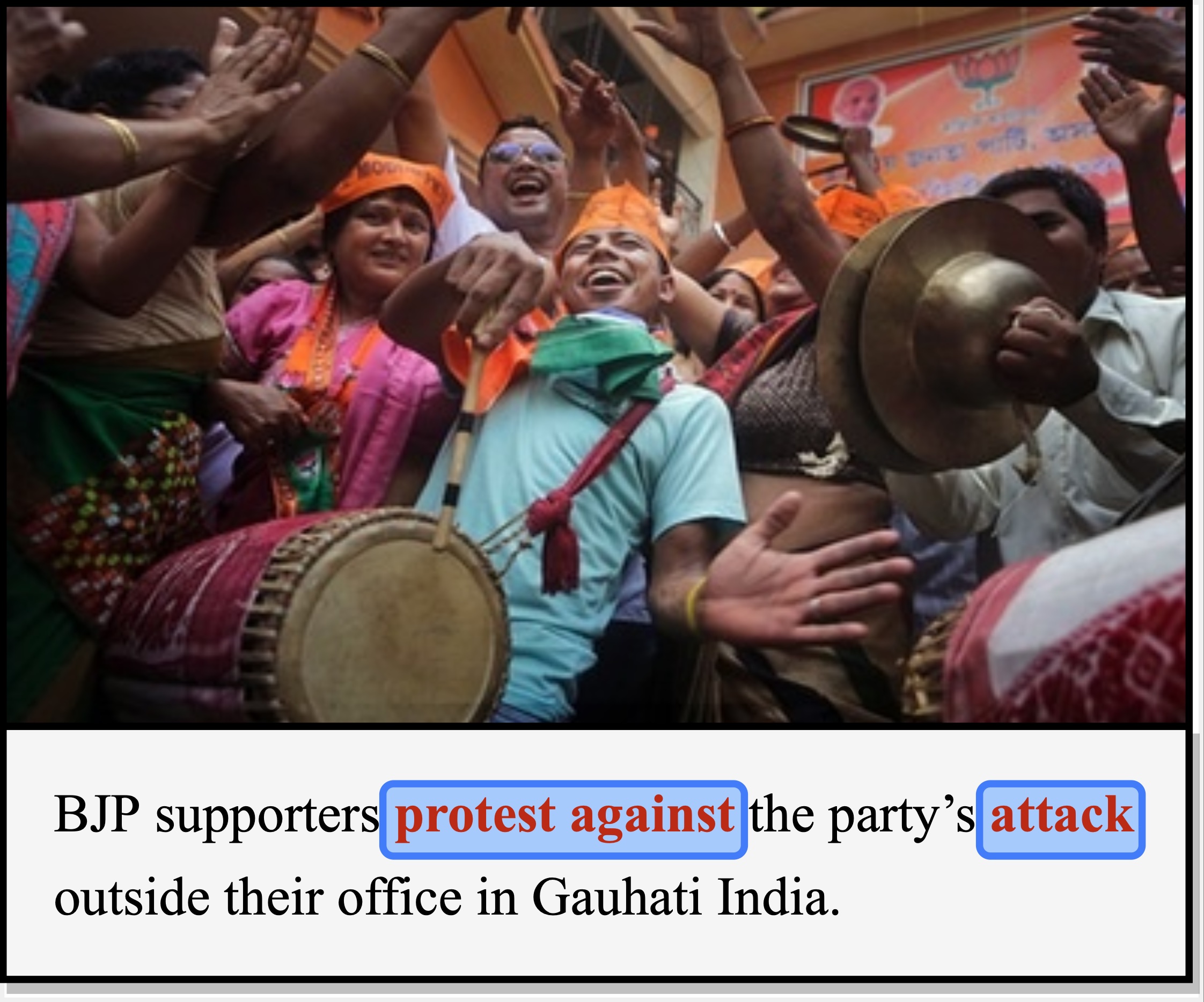}}
    \label{fig:vis_TA}
    \caption{Visualization of detection and grounding results. Ground truth annotations are in red, and prediction results are in blue.}
    \label{fig:visualization}
\end{figure}

\begin{figure}[t]
    \centering
    \subfloat[Attention map in TS]{\includegraphics[width = 0.24\textwidth]{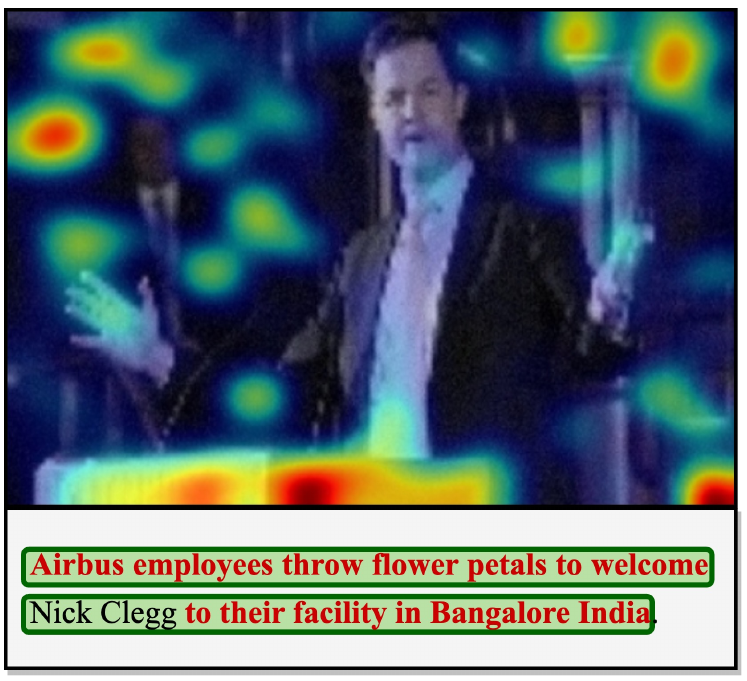}}
    \label{fig:attn_TS}
    \hfill
    \subfloat[Attention map in TA]{\includegraphics[width = 0.24\textwidth]{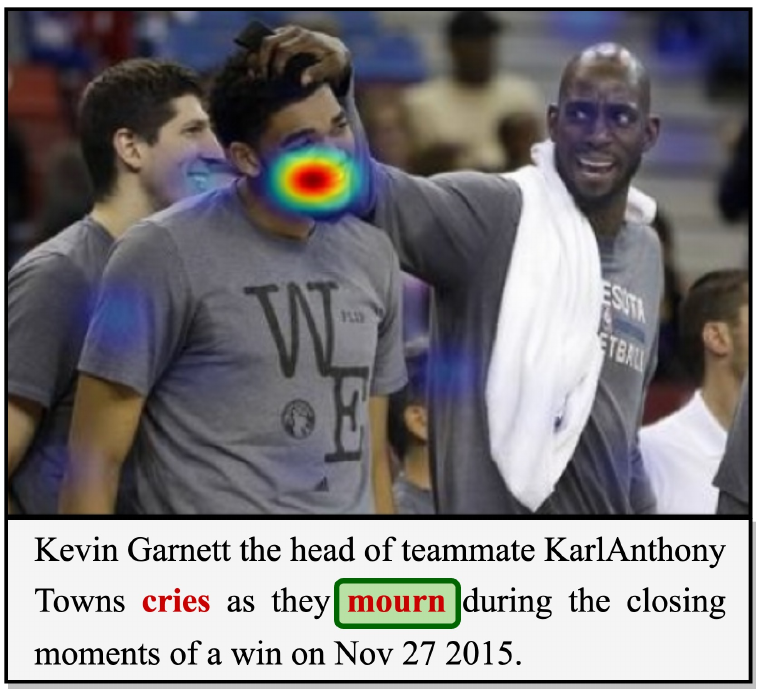}}
    \label{fig:attn_TA}
    \caption{Visualization of detection and grounding results. Ground truth annotations are in red, and prediction results are in blue.}
    \label{fig:attnmap}
\end{figure}
\subsection{Experimental Analysis}
\noindent \textbf{Ablation study of two modalities.} To validate the importance of multi-modal correlation for our model, we perform ablation study by only keeping corresponding input and network components with respect to image (Ours-Image) or text (Ours-Text) modality. \textcolor{black}{The ablated model in two modalities are depicted in Fig.~\ref{fig:ablation_study}.} We tabulate results in Tables~\ref{tab:abl_img} and~\ref{tab:abl_txt}, showing the performance of complete version of our model surpasses its ablated parts, especially in text modality. This suggests the performance degrades once one of the two modalities is missing without cross-modal interaction. This is to say, by exploiting the correlation between two modalities via our model, more complementary information between them can be dug out to promote our task. This correlation is particularly more essential for manipulation detection and grounding in the text modality.

\noindent \textbf{Ablation study of losses.} 
The considered losses and corresponding results obtained for each case are tabulated in Table~\ref{tab:abl_loss}. It is evident that removing the task-general loss, \textit{i.e.,} $\mathcal{L}_{MAC-G}$, results in nearly all the performance degeneration. This implies manipulation-aware contrastive learning is indispensable for our task. After getting rid of any one of task-specific losses, \textit{i.e.,} $\mathcal{L}_{MLC}$, $\mathcal{L}_{IMG}$ and $\mathcal{L}_{TMG}$, not only the performance of the corresponding task degrades dramatically, but also the overall binary classification performance probably becomes lower. Comparatively, our model with the complete loss function obtains the best performance in most of cases, indicating the effectiveness and complementarity of all losses. In particular, the first row of Table~\ref{tab:abl_loss} represents the current multi-modal misinformation detection scenario where only $\mathcal{L}_{BIC}$ is used. Our method substantially outperforms this baseline on binary classification, implying more manipulation grounding tasks in \textbf{DGM$^{4}$} facilitate binary classification as well. \textcolor{black}{Additionally, as shown in the last row of Table~\ref{tab:abl_loss}, adding Manipulation-Aware Contrastive Loss with Local View ($\mathcal{L}_{MAC-L}$) further benefits most of cases, which demonstrates the effectiveness of this loss.}

\noindent \textbf{Efficacy of LPAA.} 
Regarding manipulated bbox grounding, we compare the usage of \texttt{[CLS]} token~\cite{zeng2021multi} with proposed LPAA in Fig.~\ref{fig:abl_lpaa}. It shows that LPAA yields better performance under all metrics, verifying performing the manipulated bbox grounding based on attentionally aggregated patch tokens is more effective than \texttt{[CLS]} token.

\noindent \textbf{Details of manipulation type detection.} We plot the classification performance of each manipulation type based on the output of Multi-Label Classifier in Fig.~\ref{fig:multicls}. The results deliver more interpretation that text manipulation detection is harder than image modality and TA is the hardest case.

\begin{table}[t]
\scriptsize
\centering
\caption{\textcolor{black}{Generalization ability to unseen manipulations.}}
\textcolor{black}{\begin{tabular}{c|ccc}
\toprule[1pt]
Evaluation on FaceForensics++~\cite{rossler2019faceforensics++}  & \multicolumn{3}{c}{Real/Fake Cls}                                                                \\ \hline
Methods       & AUC            & EER                                    & ACC                                    \\ \hline
MAT~\cite{zhao2021multi}           & 60.17          & 42.08                                  & 39.73                                  \\
TS~\cite{luo2021generalizing}            & 61.65          & 41.93                                  & 61.69                                  \\
\rowcolor[HTML]{E3DCDC} 
\textbf{Ours} & \textbf{64.52} & \cellcolor[HTML]{E3DCDC}\textbf{39.83} & \cellcolor[HTML]{E3DCDC}\textbf{62.12} \\ \bottomrule[1pt]
\end{tabular}}
\label{tab:Generalization}
\end{table}

\noindent \textcolor{black}{\noindent \textbf{Effectiveness of manipulation-aware contrastive learning.} We compare the proposed Manipulation-Aware Contrastive Learning with normal contrastive learning in Table~\ref{tab:abl_manipulation_aware}. It can be observed that Manipulation-Aware Contrastive Learning with Global View (Manipulation-Aware CL-G) exceeds the normal counterpart (Normal CL) in most evaluation cases. Through further considering manipulated pairs as negative samples, manipulation-aware contrastive learning pushes manipulated pairs apart and thus emphasizes the semantic inconsistency incurred by the cross-modal manipulation, yielding better performance. Furthermore, after supplementing Manipulation-Aware Contrastive Learning with Local View (Manipulation-Aware CL-G+L), the performance of nearly all cases are further improved, demonstrating its superiority and complementarity compared to normal contrastive learning.}

\noindent \textcolor{black}{\noindent \textbf{Ablation study of momentum models.}
In this ablation study, we remove momentum models of two uni-modal encoders, Multi-Modal Aggregator and Token Detector. The experimental results are reported in Table~\ref{tab:abl_mom} denoted as Ours w/o Mom. As evident from Table~\ref{tab:abl_mom}, the performance of overall binary classification and manipulated token grounding degenerates without momentum models. This suggests that momentum models facilitate the contrastive learning between two uni-modals, and enable Token Detector to more robustly and accurately ground manipulated tokens facing noisy texts.}

\noindent \textcolor{black}{\noindent \textbf{Generalization ability to unseen manipulations.} In this section, we further evaluate the generalization ability of our method to unseen manipulations. To this end, we test our method to detect the deepfake data produced by unseen manipulation approaches from the publicly available dataset FaceForensics++ (FF++)~\cite{rossler2019faceforensics++}. We tabulate the comparison results with some representative deepfake detection competitors in Table~\ref{tab:Generalization}. As can be seen in this table, the proposed method also surpasses all the other competitors, suggesting its better generalization ability in the cross-dataset testing scenario.}

\begin{figure*}
	\begin{center}
		\includegraphics[width=0.92\linewidth]{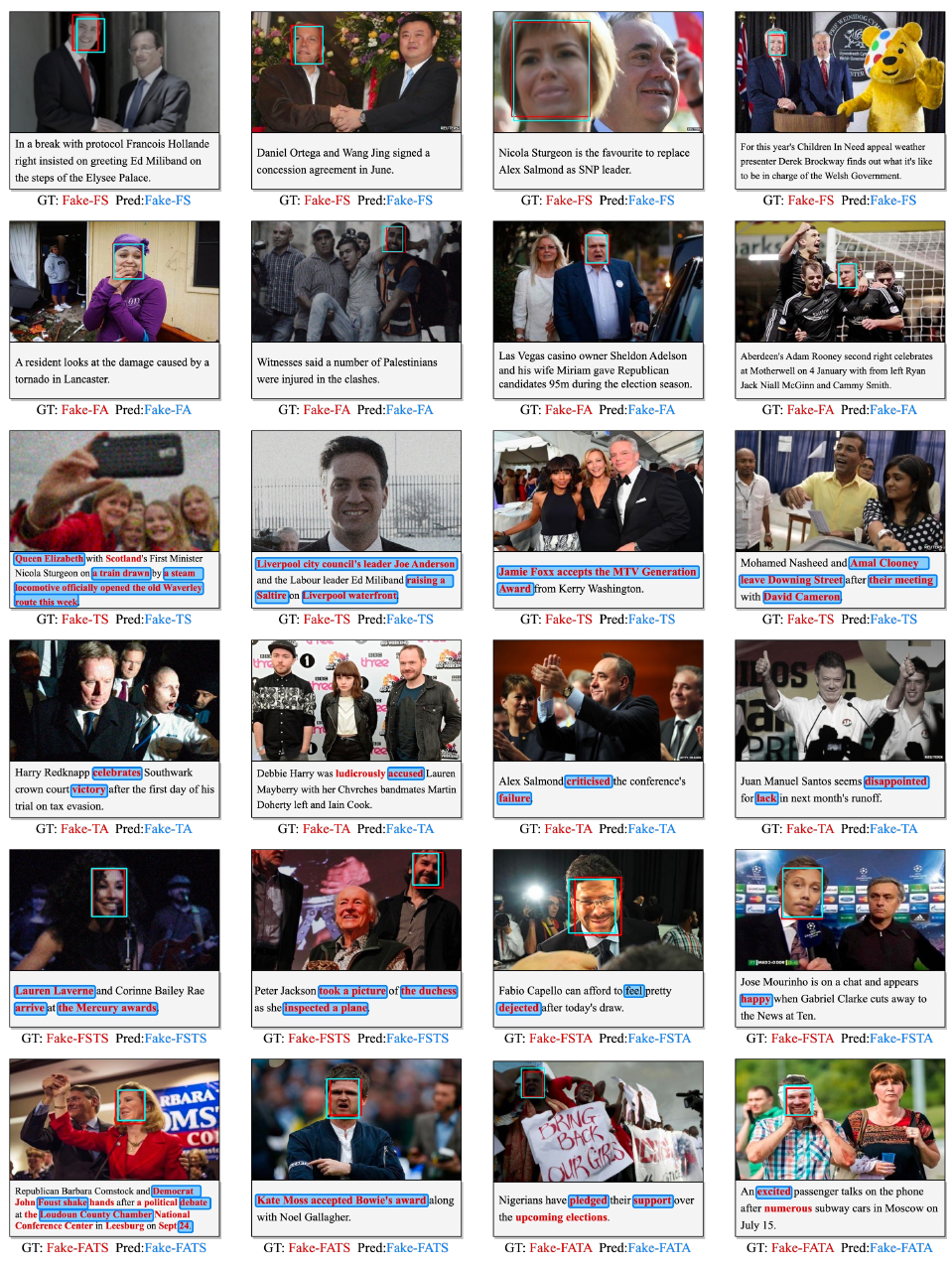}
	\end{center}
        \vspace*{-3mm}
	\caption{\textcolor{black}{More detection and grounding results. Ground truth annotations (manipulation type, manipulation bbox and manipulated tokens) are in red, prediction results of our model are in blue.}}
	\label{fig:supp_success}
\end{figure*}

\begin{figure*}
    \centering
    \subfloat[Attention map in FS]{\includegraphics[width = 0.84\textwidth]{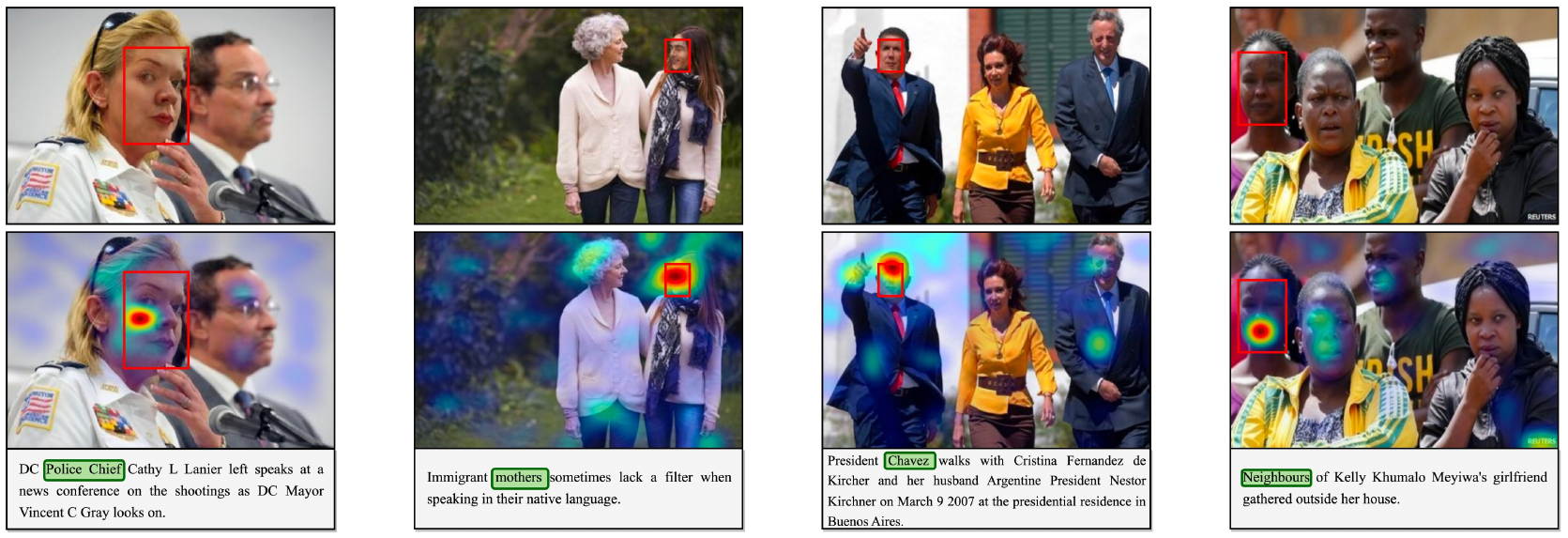}}
    \label{fig:attn_FS_supp}
    \hfill
    \subfloat[Attention map in FA]{\includegraphics[width = 0.84\textwidth]{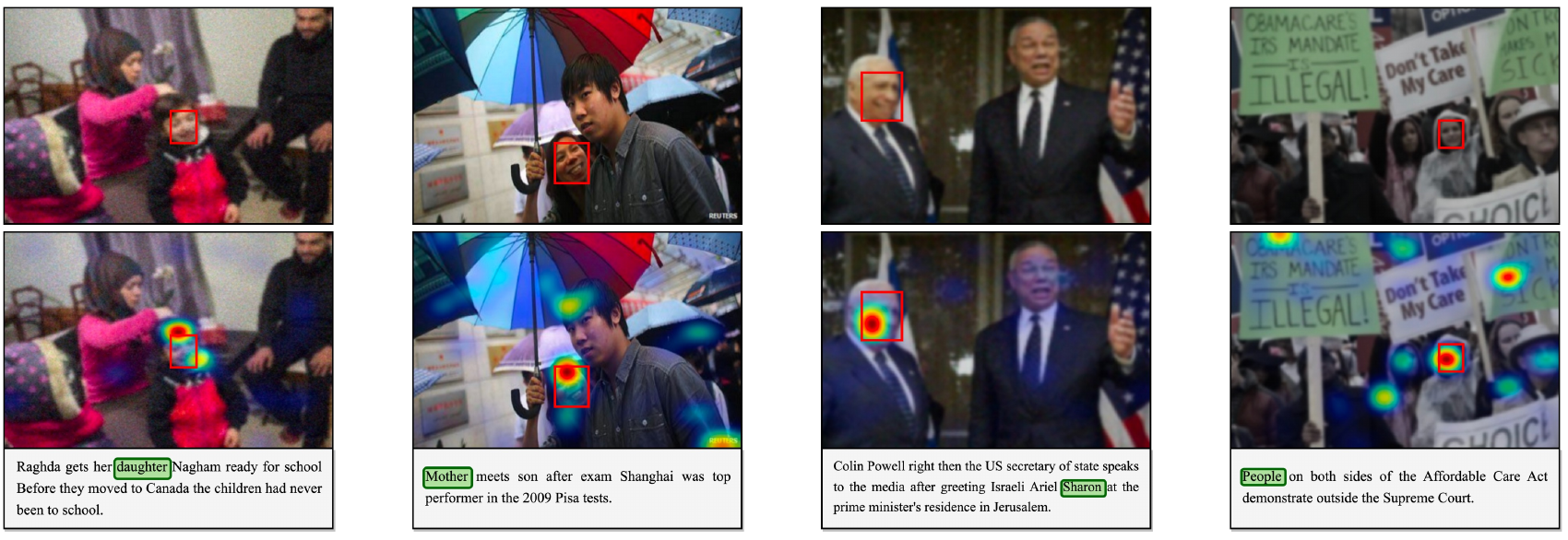}}
    \label{fig:attn_FA_supp}
    \subfloat[Attention map in TS]{\includegraphics[width = 0.84\textwidth]{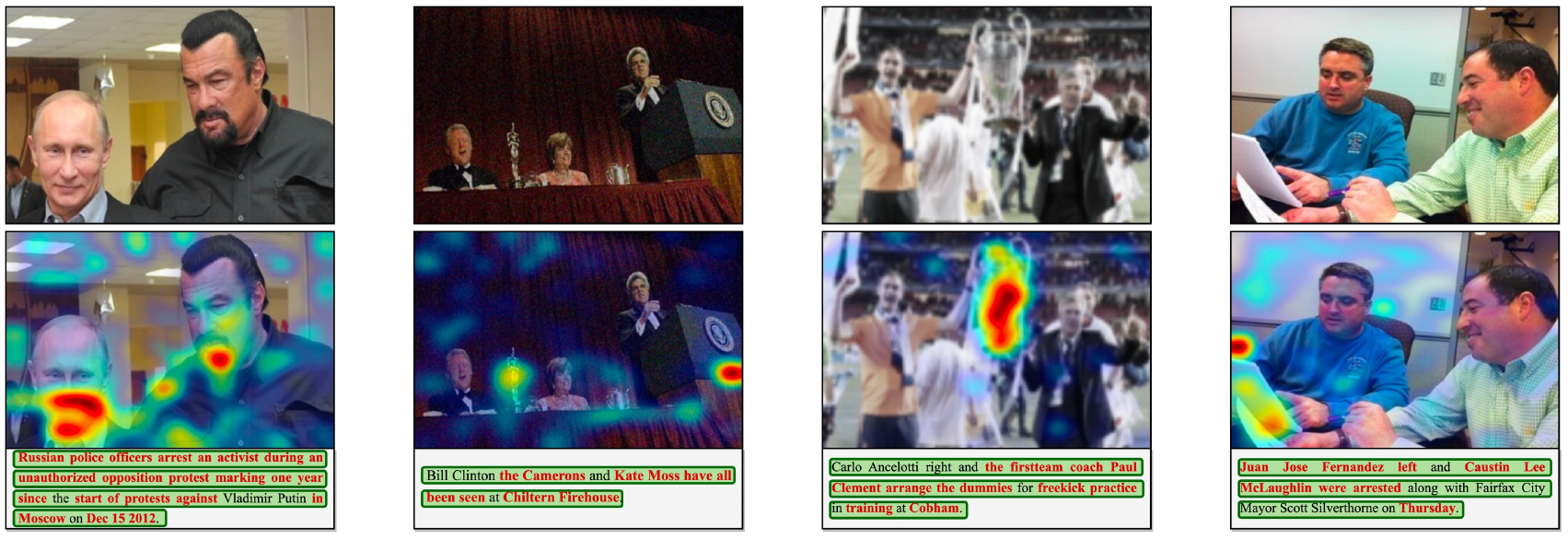}}
    \label{fig:attn_TS_supp}
    \hfill
    \subfloat[Attention map in TA]{\includegraphics[width = 0.84\textwidth]{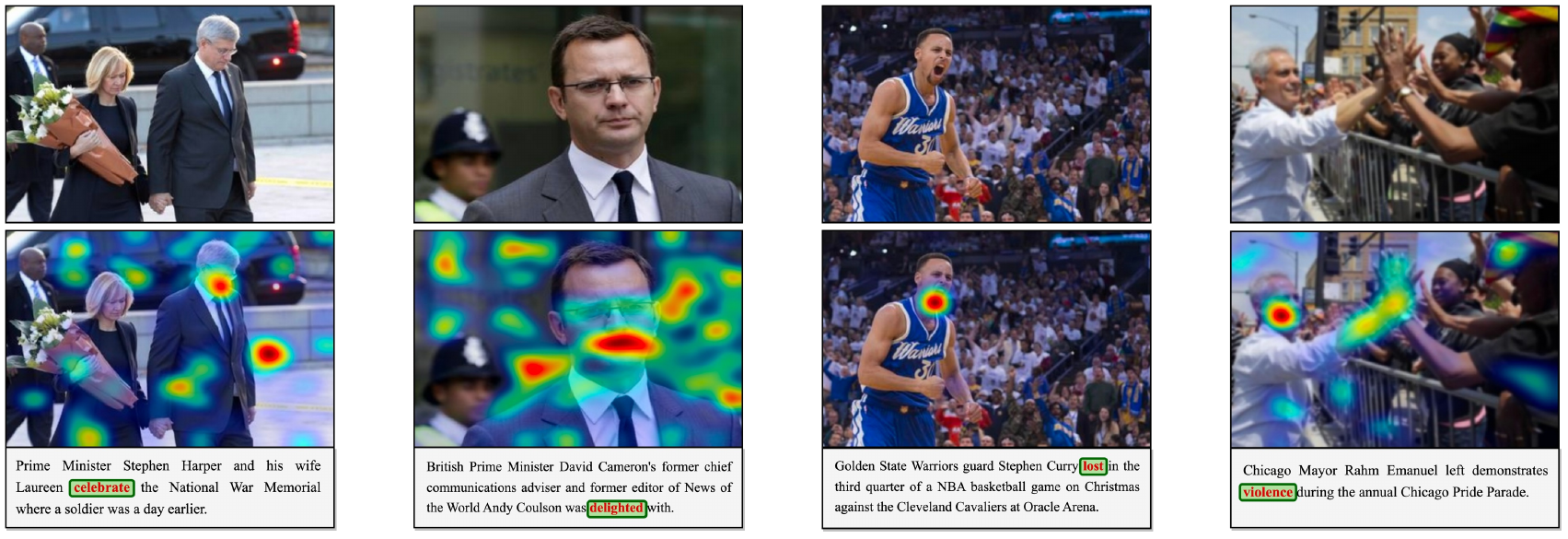}}
    \label{fig:attn_TA_supp}
    \caption{\textcolor{black}{Grad-CAM visualizations with respect to specific text tokens (in green) for four manipulated types.}}
    \label{fig:attnmap_supp}
\end{figure*}

\begin{figure*}[t] 
	\begin{center}
		\includegraphics[width=0.95\linewidth]{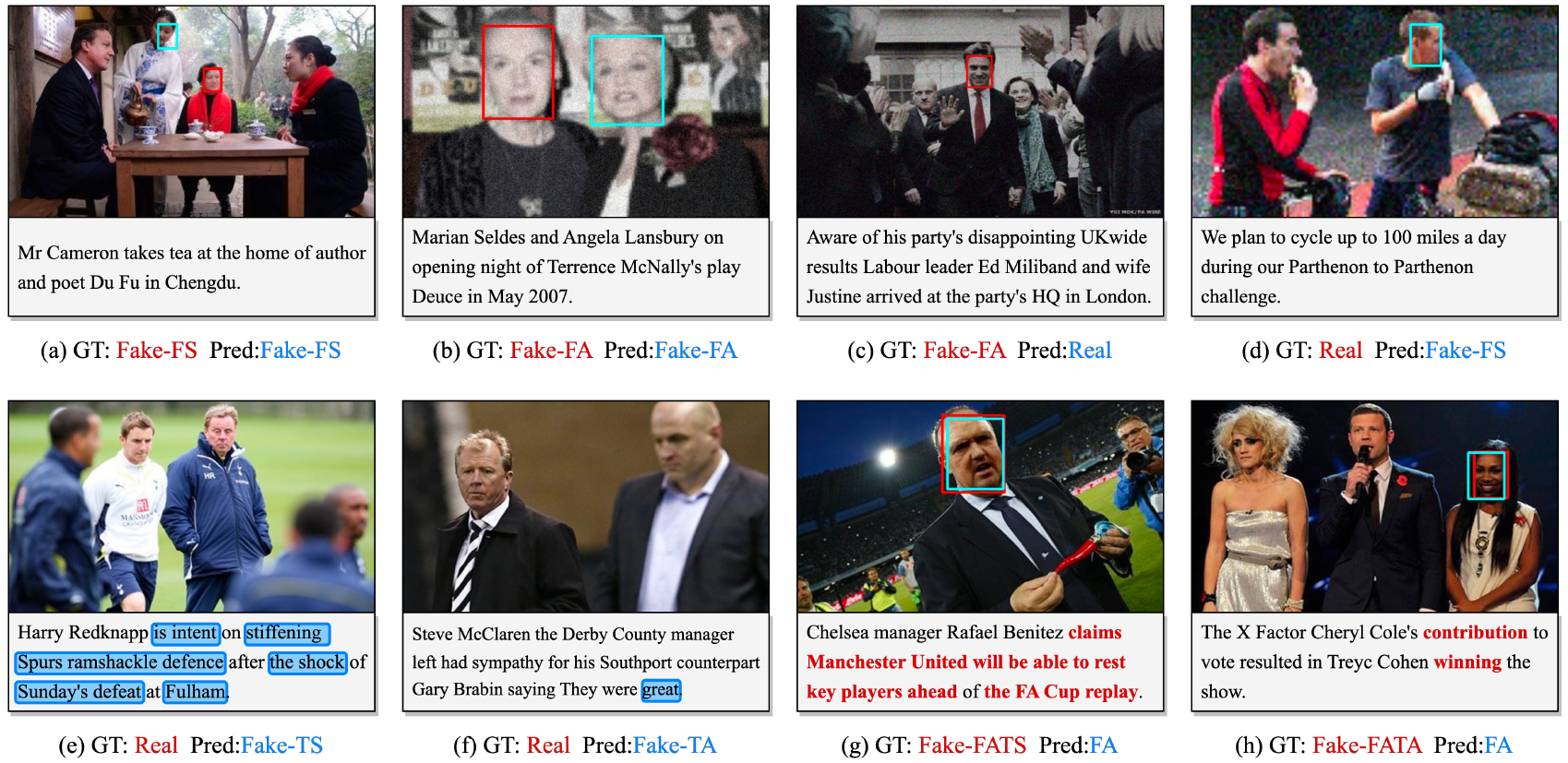}
	\end{center}
	\caption{\textcolor{black}{Failure cases of several manipulation types. They demonstrate the challenge of \textbf{DGM$^4$} and motivate us to continually improve the performance of the proposed \textbf{HAMMER} in the future.}}
	\label{fig:supp_failure}
\end{figure*}
\subsection{Visualizations}
\noindent \textbf{Visualization of manipulation detection and grounding.}
We provide some visualized results of manipulation detection and grounding in Fig.~\ref{fig:visualization}. Fig.~\ref{fig:visualization} (a)-(b) show our method can accurately ground manipulated bboxes and detect correct manipulation types for both FA and FS. Furthermore, most of the manipulated text tokens in TS and all of those in TA are successfully grounded in Fig.~\ref{fig:visualization} (c)-(d). \textcolor{black}{More visualized detection and grounding results are displayed in Fig.~\ref{fig:supp_success}. All of them visually verify that our method can achieve both effective manipulation detection and grounding.}

\noindent \textbf{Visualization of attention map.}
We provide Grad-CAM visualizations of our model regarding manipulated text tokens in Fig.~\ref{fig:attnmap}. Fig.~\ref{fig:attnmap} (a) shows our model pays attention to surroundings of the character in image. These surroundings indicate the character is giving a speech, which is semantically distinct from text tokens manipulated by TS. As for TA, Fig.~\ref{fig:attnmap} (b) shows the per-word visualization with respect to the manipulated word (`mourn'). It implies our model focuses on the smiling face in image that is semantically inconsistent with the sad sentiment expressed from the manipulated word (`mourn'). These samples prove our model indeed captures the semantic inconsistency between images and texts to tackle \textbf{DGM$^4$}. 

\textcolor{black}{We plot more Grad-CAM visualizations with respect to some specific text tokens (in green) for all the four manipulated types (\textit{i.e.,} FS, FA, TS, TA) in Fig.~\ref{fig:attnmap_supp}. For FS and FA, we visualize the attention map regarding some key words related to image manipulation. As depicted in Fig.~\ref{fig:attnmap_supp} (a)-(b), our model can accurately focus on manipulated faces based on mismatched words. For example, as can be seen from the third column in Fig.~\ref{fig:attnmap_supp} (a)-(b), our model is able to locate manipulated faces based on corresponding celebrates' names such as `Chavez' and `Sharon'. This means the proposed model can effectively unveil the inconsistent correspondence between manipulated faces and text words for image manipulation grounding. Meanwhile, it can be observed from other columns in Fig.~\ref{fig:attnmap_supp} (a)-(b) our model can also correctly attend to swapped or edited faces based on some general terms, \textit{e.g.,} `police chief', `mothers', `neighbours', `daughter' and `people'. These visualizations demonstrate text tokens can be exploited to facilitate locating manipulated image regions by our model as well.}

\textcolor{black}{Furthermore, we provide the Grad-CAM visualizations in terms of swapped texts in Fig.~\ref{fig:attnmap_supp} (c). It can be seen that the proposed model could be aware of semantically inconsistent signs in image regions compared to manipulated texts. To be specific, as illustrated in the first column of Fig.~\ref{fig:attnmap_supp} (c), manipulated text tokens talk about arresting an activist during protests about Putin. Our model could find the semantic conflict between images and manipulated text tokens as it observes the smiling face of Putin and his normal relation to the person on his right. Besides, it is evident from the second column of Fig.~\ref{fig:attnmap_supp} (c) that our model notices the surroundings around the characters are related to a banquet or ceremony, which conflicts with `Firehouse' mentioned in the text; the third column of Fig.~\ref{fig:attnmap_supp} (c)
suggests that our model focuses on the champion cup. This indicates our model understands there appears to be a celebration in image instead of the training scenario discussed in text. Furthermore, Fig.~\ref{fig:attnmap_supp} (d) provides a few per-word visualizations about manipulated tokens in TA. As evident from Fig.~\ref{fig:attnmap_supp} (d), the proposed model can pay attention to the emotions in faces that are mismatched to the corresponding semantics expressed in the manipulated words. For instance, the terms `celebrate' and `delighted' mismatch the sad emotions in faces. Meanwhile, the fourth column of Fig.~\ref{fig:attnmap_supp} (d) demonstrates our model can perceive the action of a high five in image that is contradictory to the `violence' in the text. All of the above visualization samples imply that our model can indeed capture the semantic inconsistency between images and texts to tackle \textbf{DGM$^4$}.}\\
\noindent \textcolor{black}{\noindent \textbf{Failure cases.} It can be observed from Fig.~\ref{fig:supp_failure} (a)-(d) that when the input image is severely corrupted by perturbation or the manipulated face is too small, detection and grounding tasks related to image modality may fail, resulting in false negatives and false positives. Fig.~\ref{fig:supp_failure} (e)-(f) visualize another representative failure case where the real image and text are originally not well-aligned in semantics or sentiment. This would cause the model to predict false manipulated tokens in text. We observe that this situation often occurs when the text contains the main character's comments, intentions, thoughts or other abstract concepts that are invariant to image contents. Fig.~\ref{fig:supp_failure} (g)-(h) show that mixed-manipulation pairs could be more challenging to be detected and grounded. Specifically, in Fig.~\ref{fig:supp_failure} (h), the TA-manipulated text tokens originally contradict the image content in sentiment. But after FA manipulation, the contradiction is partially eliminated, which makes the forgeries substantially harder to be detected. These failure cases indicate the difficulties of \textbf{DGM$^4$} and motivate us to continually improve our model in the future.}\\
\section{Conclusion}
\label{sec:conclusion}
This paper studies a novel \textbf{DGM$^4$} problem, aiming to detect and ground multi-modal manipulations. We construct the first large-scale \textbf{DGM$^4$} dataset with rich annotations. The powerful models \textbf{HAMMER} and \textbf{HAMMER++} are proposed and extensive experiments are performed to demonstrate their effectiveness. Various valuable insights are also provided to facilitate future research in multi-modal media manipulation.

\noindent \textbf{Potential Negative Impact.} Although image and text forgery data is generated in \textbf{DGM$^4$} dataset, the purpose is to develop a manipulation detection and grounding method to help people better fight against the abuse of multi-modal forgery. Our study hopes to draw greater research and societal attention to multi-modal media manipulation.

\ifCLASSOPTIONcaptionsoff
  \newpage
\fi

\bibliographystyle{IEEEtran}
\bibliography{ref}

\end{document}